\documentclass{article}

\usepackage{arxiv}

\usepackage{mathtools}

\usepackage[utf8]{inputenc} 
\usepackage[T1]{fontenc}  
\usepackage{hyperref} 
\usepackage{url}   
\usepackage{booktabs} 
\usepackage{amsfonts} 
\usepackage{nicefrac}  
\usepackage{microtype} 
\usepackage{lipsum}
\usepackage{graphicx}
\usepackage[graphicx]{realboxes}
\graphicspath{ {./images/} }

\usepackage{tabularx}
\usepackage{array,multirow}
\newcolumntype{K}{>{\centering\arraybackslash}m{3cm}}
\newcolumntype{L}{>{\centering\arraybackslash}m{1.3cm}}
\newcolumntype{M}{>{\centering\arraybackslash}m{1.5cm}}
\newcolumntype{N}{>{\centering\arraybackslash}m{0.8cm}}
\newcolumntype{O}{>{\centering\arraybackslash}m{2cm}}
\newcolumntype{P}{>{\centering\arraybackslash}m{0.5cm}}
\newcolumntype{B}{>{\centering\arraybackslash}m{0.35cm}}
\newcolumntype{A}{>{\centering\arraybackslash}m{0.7cm}}
\newcolumntype{Q}{>{\arraybackslash}m{9cm}}
\newcolumntype{R}{>{\arraybackslash}m{2cm}}
\newcolumntype{W}{>{\centering\arraybackslash}m{1.1cm}}
\newcolumntype{Y}{>{\raggedright\arraybackslash}X}
\newcolumntype{?}{!{\vrule width 1pt}}
\usepackage{subfig}
\usepackage[numbers]{natbib}
\usepackage{flexisym}
\usepackage{textcomp}
\usepackage{amssymb}
\usepackage{xcolor}
\usepackage{geometry}
\usepackage{makecell}
\usepackage{color,soul}
\usepackage{textcomp}
\usepackage{color, colortbl}
\definecolor{Gray}{gray}{0.9}
\usepackage{wrapfig}
\usepackage{lscape}
\usepackage{rotating} 
\usepackage{tablefootnote}
\usepackage{setspace}
\usepackage{amsmath}
\usepackage{diagbox}
\usepackage{pdflscape}

\newtheorem{proposition}{Proposition}[section]
\newtheorem{definition}{Definition}[section]

\usepackage{pifont}

\newcommand*{\TakeFourierOrnament}[1]{{%
\fontencoding{U}\fontfamily{futs}\selectfont\char#1}}
\newcommand*{\danger}{\TakeFourierOrnament{66}}

\usepackage{lineno,hyperref}
\modulolinenumbers[5]

\title{Machine learning fairness notions: Bridging the gap with real-world applications}

\author{
 Karima Makhlouf \\
  Université du Québec à Montréal\\
  Québec, Canada \\
  \texttt{karima.makhlouf@courrier.uqam.ca} \\
   \And
 Sami Zhioua \\
  Higher Colleges of Technology\\
  Dubai, United Arab Emirates \\
  \texttt{szhioua@hct.ac.ae} \\
  \And
 Catuscia Palamidessi \\
  Inria, Ecole Polytechnique, IPP\\
  Paris, France \\
  \texttt{catuscia@lix.polytechnique.fr} \\
  }

\begin{document}
\maketitle
\begin{abstract}
Fairness emerged as an important requirement to guarantee that Machine Learning (ML) predictive systems do not discriminate against specific individuals or entire sub-populations, in particular, minorities. Given the inherent subjectivity of viewing the concept of fairness, several notions of fairness have been introduced in the literature. This paper is a survey  that illustrates the subtleties between fairness notions through a large number of examples and scenarios. In addition, unlike other surveys in the literature, it addresses the question of ``which notion of fairness is most suited to a given real-world scenario and why?''. 
Our attempt to answer this question consists in (1) identifying the set of fairness-related characteristics of the real-world scenario at hand, (2) analyzing the behavior of each fairness notion, and then (3) fitting these two elements to recommend the most suitable fairness notion in every specific setup. The results are summarized in a decision diagram that can be used by practitioners and policy makers to navigate the relatively large catalogue of ML fairness notions.
\end{abstract}



\section{Introduction}
\label{intro}
Decisions in several domains are increasingly taken by ``machines''. These machines try to take the best decisions based on relevant historical data and using Machine Learning (ML) algorithms. Overall, ML-based decision-making (MLDM)\footnote{We focus on automated decision-making system supported by ML algorithms. In the rest of the paper we refer to such systems as MLDM.} is  beneficial as it allows to take into consideration orders of magnitude more factors than humans do and hence outputting decisions that are more informed and less subjective. However, in their quest to maximize efficiency, ML algorithms can systemize discrimination against a specific group of population, typically, minorities. As an example, consider the automated candidates selection system of St. George Hospital Medical School~\citep{george88,weapons16}. The aim of the system was to help screening for the most promising candidates for medical studies. The automated system was built using records of manual screenings from previous years. During those manual screening years, applications with grammatical mistakes and misspellings were rejected by human evaluators as they indicate a poor level of English. As non-native English speakers are more likely to send applications with grammatical and misspelling mistakes than native English speakers do, the automated screening system built on that historical data ended up correlating race, birthplace, and address with a lower likelihood of acceptance. Later, while the overall English level of non-native speakers improved, the race and ethnicity bias persisted in the system to the extent that an excellent candidate may be rejected simply for her birthplace or address.

Given that  MLDM can have a significant impact in the lives and safety of human beings, it is no surprise that social and political organization 
are becoming very  concerned with the possible consequences of biased MLDM, and the related issue of lack of explanation and interpretability of   ML-based decisions. 
The European Union has been quite active in this respect:  already in the General Data Protection Regulation (GDPR) there were directives concerning Automated Decision Making: for instance, Article 22 states 
that ``The data subject shall have the right not to be subject to a decision based solely on automated processing.''
Other initiatives include the European Union’s Ethics Guidelines for Trustworthy AI (April 2019), and OECD’s Council Recommendation
on Artificial Intelligence (May 2019). 

In the scientific community, the issue of fairness in machine learning has become one of the most popular topics  in recent years. The number of 
publications and conferences in this field has literally exploded, and a huge number of different notions of fairness have been proposed, leading sometimes to possible confusion. 
This paper, like other surveys in the literature (cf. Section~\ref{sec:relatedWork}), attempts to classify and systematize these notions. The characteristic of our work, however, consists in our point of view, which is that \emph{ the very reason for having different fairness notions is how suitable each one of them is for specific real-world scenarios}. We feel that none of the existing surveys has addressed this aspect specifically. Discussion about the suitability (and sometimes the applicability) of the fairness notions is very limited and scattered through several papers~\citep{mitchell2018prediction,gajane2017survey,zafar2017fairness,kleinberg_et_al:LIPIcs:2017:8156,corbett2017algorithmic,barocas-hardt-narayanan}. In this survey paper we show that each MLDM system can be different based on a set of criteria such as: whether the ground-truth exists, difference in base-rates between sub-groups, the cost of misclassification, the existence of a government regulation that needs to be enforced, etc. 
We then revisit exhaustively the list of fairness notions and discuss the suitability and applicability of each one of them based on the list of criteria. 

Another set of results from the literature which is particularly related to the applicability problem we are addressing in this paper is the tensions that exist between some definitions of fairness. Several papers in the literature provide formal proofs of the impossibility to satisfy several fairness definitions simultaneously~\citep{mitchell2018prediction,kleinberg_et_al:LIPIcs:2017:8156,barocas-hardt-narayanan,chouldechova2017fair,friedler2016possibility}. These results are revisited and summarized as they are related to the applicability of fairness notions.

The results of this survey are finally summarized in a decision diagram that hopefully can help researchers, practitioners, and policy makers to identify the subtleties of the MLDM system at hand and to choose the most appropriate fairness notion to use, or at least rule out notions that can lead to wrong fairness/discrimination result. 

The paper is organized as follows. Section~\ref{sec:scenarios} lists notable real-world MLDMs where fairness is critical. Section~\ref{sec:criteria} identifies a set of fairness-related characteristics of MLDMs that will be used in the subsequent sections to recommend and/or discourage the use of fairness notions. Fairness notions are listed and described in the longest section of the survey, Section~\ref{sec:notions}. Section~\ref{sec:relaxation} discusses relaxations of the strict definitions of fairness notions. Section~\ref{sec:trade-offs} describes classification and tensions that exist between some fairness notions. The decision diagram is provided and discussed in Section~\ref{sec:diagram}. 

\color{black}
\section{Related Work and Scope}
\label{sec:relatedWork}

With the increasing fairness concerns in the field of automated decision making and machine learning, several survey papers have been published in the literature in the few previous years. This section revisits these survey papers and highlights how this proposed survey deviates from them. 

In 2015, Zliobaite compiled a survey about fairness notions that have been introduced previously~\citep{zliobaite2015survey}. He classified fairness notions into four categories, namely, statistical tests, absolute measures, conditional measures, and structural measures. Statistical tests indicate only the presence or absence of discrimination. Absolute and conditional measures quantify the extent of discrimination with the difference that conditional measures consider legitimate explanations for the discrimination. These three categories correspond to the group fairness notions in this survey. Structural measures correspond to individual fairness notions\footnote{Zliobaite does not use group vs individual notions, but indirect and direct discrimination.}. Most of the fairness notions listed by Zliobaite are variants of the group fairness notions in this survey. For instance, difference of means test (Section 4.1.2 in~\citep{zliobaite2015survey}) is a variant of balance for positive class (Section~\ref{sec:balance} in this paper). Although, he dedicated one category for individual notions (structural measures), Zliobaite did not mention important notions, in particular fairness through awareness. Regarding the applicability of notions, the only criterion considered was the type of variables (e.g. binary, categorical, numerical, etc.). 

The survey of Berk et al.~\citep{berk2018fairness} listed only group fairness notions that are defined using the confusion matrix. Similar to this survey, they used simple examples based on the confusion matrix to highlight relationships between the fairness notions. The applicability aspect has not been addressed as the paper focused only on criminal risk assessment use case.

The survey of Verma and Rubin~\citep{verma2018fairness} described a list of fairness notions similar to the list in this survey. To illustrate how each notion can be computed in real scenarios, they used a loan granting real use case (German credit dataset~\citep{uci2007}). Rather than using a benchmark dataset, this survey uses a smaller and fictitious use case (job hiring) which allows to illustrate better the subtle differences between the fairness notions. For instance, counterfactual fairness is more intuitively described using a small job hiring example than the loan granting benchmark dataset. Verma and Rubin did not address the applicability aspect in their survey.

\textcolor{black}{
Gajane and Pechenizkiy~\cite{gajane2017survey} focused on formalizing only notable fairness notions (e.g. statistical parity, equality of opportunity, individual fairness, etc.) and discussed their implications on distributive justice from the social sciences literature. In addition, they described two additional fairness notions that are studied extensively in the social sciences literature, namely, equality of resources and equality of capability. These notions, however, do not come with a mathematical formalization. This survey is more exhaustive as it analyzes a much larger number of fairness notions. However, being focused on the implication on distributive justice, Gajane and Pechenizkiy's survey addresses the suitability of the discussed fairness notions in real world domains. }

Mehrabi et al.~\citep{mehrabi2019survey} considered a more general scope for their survey: in addition to briefly listing $10$ definitions of fairness notions (Section 4.2), they surveyed different sources of bias and different types of discrimination, they listed methods to implement fairness categorized into pre-processing, in-processing, and post-processing, and they discussed potential directions for contributions in the field. This survey is more focused on fairness notions which are described in more depth.

A more recent survey by Mitchell et al.~\citep{mitchell2018prediction} presents an exhaustive list of fairness notions in both categories (group and individual) and summarizes most of the incompatibility results in the literature. Although Mitchell et al. discuss a ``catalogue'' of choices and assumptions in the context of fairness, the aim of these choices and assumptions is different from the criteria defined in this survey (Section~\ref{sec:criteria}). The assumptions and choices discussed in Section~2 in \citep{mitchell2018prediction} address the question of how social goals are abstracted and formulated into a prediction (ML) problem. In particular, how the choice of the prediction goal, the choice of the population, and the choice of the decision space can have an impact on the degree of fairness of the prediction. Whereas the choices and criteria discussed in this survey (Section~\ref{sec:criteria}) are used to help identify the most suitable fairness notion to apply in a given scenario. 

Other surveys include the one by Friedler et al.~\citep{friedler2019comparative} which considered only group fairness notions and focused on surveying algorithms to implement fairness. 

Overall most of existing review papers do not address all flavors of fairness notions in the same survey. In particular, most of them focus on statistical and group fairness notions. Causality based fairness notions, however, are not covered in several surveys while it is the most reliable category of notions in the disparate treatment legal framework.
However, the main contribution of this survey is the focus on the applicability of fairness notions and the identification of fairness-related criteria to help select the most suitable notion to use given a scenario at hand. Brief discussions about the suitability of specific fairness notions can be found in few papers. For instance, Zafar et al.~\citep{zafar2017fairness} mentioned some application scenarios for statistical parity and equalized odds. Kleinberg et al.\citep{kleinberg_et_al:LIPIcs:2017:8156} discussed the applicability of calibration and balance notions. Through a discussion about the cost of unfair decision on society, Corbett-Davies et al.\citep{corbett2017algorithmic} analyzed the impact of using statistical parity, predictive equality, and conditional statistical parity on public safety (criminal risk assessment). {\color{black} Gajane and Pechenizkiy~\cite{gajane2017survey} discuss the suitability of notable fairness notions (statistical parity, individual fairness, etc.) from the distributive justice point of view.}
Unlike the scattered discussions about the applicability of fairness notions found in the literature, this survey provides a complete reference to systemize the selection procedure of fairness notions. A short version of this paper was presented in BIAS 2020 workshop at ECMLPKDD 2020~\citep{bias2020version}.

Fairness in machine learning can be categorized according to two dimensions, namely, the task and the type of learning. For the first dimension, there are two tasks in fairness-aware machine learning: discrimination discovery (or assessment) and discrimination removal (or prevention). Discrimination discovery task focuses on assessing and measuring bias in datasets or in predictions made by the MLDM. Discrimination removal focuses on preventing discrimination by manipulating datasets (pre-processing), adjusting the MLDM (in-processing) or modifying predictions (post-processing). For the second dimension, fairness can be investigated for different learning types including fairness in classification, fairness in regression~\cite{kamishima2011fairness,agarwal2019fair}, fairness in ranking~\cite{celis2018ranking}, fairness in reinforcement learning~\cite{jabbari2017fairness}, etc. This survey focuses on the task of discrimination discovery (assessing fairness) in ``pure prediction''~\cite{kleinberg2015prediction} classification problems with a single decision making task (not sequential) and where decisions do not impact outcomes~\cite{coston2020counterfactual}. 
\\

\section{Real-world scenarios with critical fairness requirements}
\label{sec:scenarios}
As the paper is focusing on the applicability of fairness notions, we provide here a list of notable real-world MLDMs where fairness is critical. In each of these scenarios, failure to address the fairness requirement will lead to unacceptable biased decisions against individuals and/or sub-populations. These scenarios will be used to provide concrete examples of situations where certain fairness notions are more suitable than others.

\textit{\textbf{Job hiring}}: MLDMs in hiring are increasingly used by employers to automatically screen candidates for job openings\footnote{In 2014, the automated job screening systems market was estimated at \$500 million annual business and was growing at a rate of 10 to 15\% per year~\citep{jobHiringBusiness}}. Commercial candidate screening MLDMs include XING\footnote{A job platform similar to LinkedIn. It was found that this platform ranked less qualified male candidates higher than more qualified female candidates \citep{lahoti2019ifair}.}, Evolv~\citep{evolv}, Entelo, Xor, EngageTalent, GoHire \textcolor{black}{and SyRI}~\footnote{\textcolor{black}{System Riscico Indicatie, or SyRI for short, is a risk profiling system being deployed in the Netherlands by the Department of Social Affairs and Employment with the intention of identifying individuals who are at a high risk of committing fraud in relation to employment and 
other matters like social security and taxes.
Its use raised a lot of controversy, and its case was brought to the Court of 
the Hague, that   concluded on the 5th of  February 2020 that the Government’s use of SyRI violates the European Convention on Human Rights. To a very large extent, the Court’s judgment was based on the lack of transparency in the algorithm at the heart of the system.}}. 
Typically, the input data used by the MLDM include: affiliation, education level, job experience, IQ score, age, gender, marital status, address, etc. The MLDM outputs a decision and/or a score indicating how suitable/promising the application is for the job opening. A biased MLDM leads to rejecting a candidate because of a trait that she cannot control (gender, race, sexual orientation, etc.). Such unfairness causes a prejudice on the candidate but also can be damaging for the employer as excellent candidates might be missed.

\textit{\textbf{Granting loans}}: Since decades, statistical and MLDM systems are used to assess loan applications and determine which of them are approved and with which repayment plan and annual percentage rate (APR). The assessment proceeds by predicting the risk that the applicant will default on her repayment plan. Loan Granting MLDMs currently in use include:  FICO, Equifax, Lenddo, Experian, TransUnion, etc. The common input data used for loan granting include: credit history, purpose of the loan, loan amount requested, employment status, income, marital status, gender, age, address, housing status and credit score. An unfair loan granting MLDM will either deny a deserving applicant a requested loan, or give her an exorbitant APR, which on the long run will create a vicious cycle as the candidate will be very likely to default on her payments.

\textit{\textbf{College admission}}: Given the large number of admission applications, several colleges are now resorting to MLDMs to reduce processing time and cut costs\footnote{While the final acceptance decision is taken by humans, MLDMs are typically used as a first filter to ``clean-up'' the list from clear rejection cases.}. Existing college admission MLDMs include GRADE~\citep{waters2014grade}, IBM Watson\footnote{A platform that uses natural language processing and personality traits in order to help students find the suitable and right college for them.}, Kira Talent\footnote{A Canadian startup that sells a cloud-based admissions assessment platform to over $300$ schools.} . Typically, the candidates' features used include: the institutions previously attended, SAT scores, extra-curricular activities, GPAs, test scores, interview score, etc. The predicted outcome can be a simple decision (admit/reject) or a score indicating the candidate's potential performance in the requested field of study~ \citep{friedler2016possibility}. Unfair college admission MLDMs may discriminate against a certain ethnic group (e.g. African-American~\citep{santelices2010unfair}) which could lead, in the long term, to economic inequalities and corrupting the role of higher education in society as a whole.
For instance, in 2020 Ofqual, the UK Office of Qualifications and Examinations Regulation, used a MLDM to assess students for university admission decisions. Nearly 40\% of students ended up receiving exam scores downgraded from their teachers’ predictions, threatening to cost them their university spots. Analysis of the algorithm  revealed that it had disproportionately hurt students from working-class and disadvantaged communities and inflated the scores of students from private schools~\cite{UKAdmissionSystem:20}.

\textit{\textbf{Criminal risk assessment}}: 
There is an increasing adoption of MLDMs that predict risk scores based on historical data with the objective to guide human judges in their decisions. The most common use case is to predict whether a defendant will re-offend (or recidivate). Examples of risk assessment MLDMs include COMPAS~\citep{COMPAS}, PSA ~\citep{majdara2008development}, SAVRY~\citep{meyers2008predictive}, predPol~\citep{predpol}. Predicting risk and recidivism requires input information such as: number of arrests, type of crime, address, employment status, marital status, income, age, housing status, etc. Unfair risk assessment MLDMs, as revealed by the highly publicized 2016 proPublica article~\citep{angwin2016machine}, may result in biased treatment of individuals based solely on their race. In extreme cases, it may lead to wrongful imprisonments for innocent people, contributing to the cycle of violation and crime.

\textit{\textbf{Teachers evaluation and promotion}}: MLDMs are increasingly used by decision makers to decide which teachers to retain after a probationary period~\citep{chalfin2016productivity} and which tenured teachers to promote. An example of such MLDM is IMPACT~\citep{impact}. Teacher evaluation MLDMs take as input teacher related features (age, education level, experience, surveys, classroom observations), students related features (test scores, sociodemographics, surveys), and principals related features (surveys about the school and teachers), to predict whether teachers are retained. A biased teacher evaluation MLDM may lead to a systematic unfair low evaluation for teachers in poor neighborhoods, which, very often, happen to be teachers belonging to minority groups~\citep{impactBias}. On the long term, this may lead to a significant drop in students' performance and the compromise of overall school reputation~\citep{weapons16}. 

\textit{\textbf{Child maltreatment prediction}}: The objective of the MLDM in child maltreatment prediction is to estimate the likelihood of substantiated maltreatment (neglect, physical abuse, sexual abuse, or emotional maltreatment) among children. The system generates risk scores, which would then trigger a targeted early intervention in order to prevent children maltreatment. PRM (predictive risk model)~\citep{vaithianathan2013children} has been developed to estimate the likelihood of substantiated maltreatment among children enrolled in New Zealand’s public benefit system. \textcolor{black}{In Finland, the government  uses a ML-based system called ``Kela'' to administer
benefits and to identify risk factors indicating that a child might need welfare services. } In the US, the  Allegheny County uses AFST (Allegheny Family Screening Tool)~\citep{eubanks2018automating} to improve decision-making in child welfare system. The features considered in this type of MLDM include both contemporaneous and historical information for children and caregivers. An unfair MLDM may use a proxy variable to predict decisions based on the community rather than which child gets harmed. For example, a major cause of unfairness in AFST is the rate of referral calls; the community calls the child abuse hotline to report non-white families at a much higher rate than it does to report white families~\citep{eubanks2018automating}. 
On the long term, this creates a vicious cycle as families which have been reported will be the subject of more scrutiny and more requirements to satisfy, and eventually, will be more likely to fail short of these requirements and hence confirm the prediction of the system. 

\textit{\textbf{Health care}}: Since decades, ML algorithms are able to process anonymized electronic health records and flag potential emergencies, to which clinicians are invited to respond promptly. Examples of features that might be used in disease (chronic conditions) prediction include vital signs, blood test, socio-demographics, education, health insurance, home ownership, age, race, address. The outcome of the MLDM is typically an estimated likelihood of getting a disease.
A biased disease prediction MLDM can misclassify individuals in certain sub-populations in a disproportionately higher rate than the dominant population. For instance, diabetic patients have
known differences in associated complications across ethnicities \citep{spanakis2013race}. Obemeyer et al.~\citep{obermeyer2019dissecting} give another example of an MLDM that predicts the health care spending for individuals in the coming years (useful information for insurance companies). They observe that the MLDM is biased against African-Americans because it uses the cost of health services in the previous year to predict the spending in the coming years. As African-Americans were spending less on health services than whites in the previous year, they were predicted to be spending less in the coming years. Hence, for the same prediction score, African-Americans were found to be sicker (more health issues) than whites. 
Consequently, white patients were benefiting more from additional help programs than African-Americans. More generally, because different sub-populations might have different characteristics, a single model to predict complications is unlikely to be best-suited for specific groups in the population even if they are equally represented in the training data \citep{suresh2019framework}. Failure to predict disease likelihood in a timely manner may, in extreme cases, have an impact on people's lives.

\textit{\textbf{Online recommendation}}: Recommender systems are among the most widespread MLDMs in the market, with many services to assist users in finding products or information that are of potential interest \citep{jannach2010recommender}. Such systems find applications in various online platforms such as Amazon, Youtube, Netflix, LinkedIn, etc. An unfair recommender MLDM can amplify gender bias in the data. For example, a recommender MLDM called STEM, which aims to deliver advertisements promoting jobs in Science, Technology, Engineering, and Math fields, is deemed unfair as it has been shown that less women compared to men saw the advertisements due to gender imbalance \citep{lambrecht2018algorithmic}. Datta et al.~\citep{datta2015automated} found that changing the gender bit in Google Ad Setting~\citep{googleAd} resulted in a significant difference in the type of job ads received: men received much more ads about high paying jobs and career coaching services towards high paying jobs compared to women.

\textit{\textbf{Facial analysis}}: Automated facial analysis systems are used to identify perpetrators from security video footage, to detect melanoma (skin cancer) from face images~\citep{esteva2017}, to detect emotions~\citep{dehghan2017dager, fabian2016emotionet, srinivasan2016neural}, and to even determine individual's characteristics such as IQ, propensity towards terrorist crime, etc. based on their face images~\citep{wu2016automated}. \textcolor{black}{The possible applications of Facial Analysis are innumerable. 
For instance, in France, FRT (Facial Recognition  Tool) has been used on an experimental basis at various schools, with the aim of making access more fluid and secure for
pupils. Furthermore, the government announced in 2020 that it would start to use an FRT system called “Alicem” in order to create a digital identification system by which its citizens
could access governmental online services. Both of these, however, have sparked a lot of controversy
leading to an announcement that the French government would be
reviewing the use of FRT.
Indeed, these devices are particularly intrusive and present major risks of invasion of the privacy and individual freedoms. Worse yet,} a flawed MLDM may lead to biased outcomes such as wrongfully accusing individuals from specific ethnic groups (e.g. Asians, dark skin populations) for crimes (based on security video footage) at a much higher rate than the rest of the population. For instance, African-Americans have been reported to be more likely to be stopped and investigated by law enforcement due to a flawed face recognition system~\citep{garvie2016perpetual}. An investigation of three commercial face-based gender classification systems found that the error rate for dark-skinned females can be as high as $34.7\%$ while for light-skinned males the maximum error rate is $0.8\%$~\citep{buolamwini2018gender}. 

\textit{\textbf{Others}}: Other MLDMs with fairness concerns include: insurance policy prediction~
\citep{shrestha2019fairness}, income prediction~\citep{mehrabi2019survey}, \citep{zhao2018employee, esmaieeli2015data, sexton2005employee,alao2013analyzing}, and university ranking~\citep{marope2013rankings,weapons16}. 

\textcolor{black}{
For a survey of the various kinds of MLDMs  used in European countries,  
and a description of the debates and legal actions they have triggered, we recommend the excellent report by Robin Allen QC and Dee Masters~\cite{AllenDC:20:EQUINET} for the   European Network of Equality Bodies.}

\section{Fairness notion selection criteria}\label{sec:criteria}
In order to systemize the procedure for selecting the most suitable fairness notion for a specific MLDM system, we identify a set of criteria that can be used as as roadmap. For each criterion, we check whether it holds in the problem at hand or not. Telling whether a criterion is satisfied or not does not typically require an expertise in the problem domain. 

{\color{black}This section presents a list of 13 selection criteria. These criteria are derived mainly from three sources. First, the types of bias. For instance, the unreliable outcome criterion is a manifestation of a historical bias. Second, the mathematical formulation of the fairness notions themselves. For instance, the emphasis on precision vs recall criterion reflects a fundamental difference in the mathematical formulations of two families of notions, namely, predictive parity and equal opportunity. Third, the existing anti-discrimination legislation. The last two criteria are inspired by the current legislation.} 

We note here that in some cases, these criteria can, not only indicate if a fairness notion is suitable, but whether it is ``acceptable'' to use in the first place. 

\textit{\textbf{Ground truth availability}}: A ground truth value is the true and correct \textit{observed} outcome corresponding to given sample in the data. It should be distinguished from an \textit{inferred} subjective outcome in historical data which is decided by a human.  An example of a scenario where ground truth is available is when predicting whether an individual has a disease. The ground truth value is observed by submitting the individual to a blood test\footnote{Assuming the blood test is flawless.} for example. An example of a scenario where ground truth is not available is predicting whether a job applicant is hired. The outcome in the training data is inferred by a human decision maker which is often a subjective decision, no matter how hard she is trying to be objective. It is important to mention here that the availability of the ground truth depends on how the outcome is defined. Consider, for example, college admission scenario. If the outcome in the training data is defined as whether the applicant is admitted or rejected, ground truth is not available. If, however, the outcome is defined as whether the applicant will ultimately graduate from college with a high GPA, ground truth is available as it can be observed after a couple of years.

\textit{\textbf{Base rate is the same across groups}}: The base rate is the proportion of positive outcome in a population (Table~\ref{tab:confMat}). A positive outcome is the goal of the prediction (e.g. a candidate to college is admitted, a child is maltreated, an individual is granted a loan, etc.). Note that the positive outcome can be desirable (e.g. hiring, admission) or undesirable (e.g. firing, high criminal risk). The base rate can be the same or differs across sub-populations. For example, the base rates for diabetes disease occurrence for men and women is typically the same. But, for another disease such as prostate cancer, the base rates are different between men and women\footnote{While male prostate cancer is the second most common cancer in men, female prostate cancer is rare~\citep{dodson1994skene}.}.

\textit{\textbf{(Un)reliable outcome}}: In scenarios where ground truth is not available, the outcome (label) in the data is typically inferred by humans. The outcome in the training data in that case can or cannot be reliable as it can encode human bias. The reliability of the outcome depends on the data collection procedure and how rigorous the data has been checked. Scenarios such as job hiring and college admission may be more prone to the unreliable outcome problem than recommender system for example. A ``one-size-fit-all'' MLDM model in disease prediction that does not take into consideration the ethnic group of the individual may result in unreliable outcome as well.

\textit{\textbf{Presence of explanatory} variables}: An explanatory variable\footnote{Referred also as a resolving variable.} is correlated with the sensitive attribute (e.g. race) in a legitimate way. Any discrimination that can be explained using that variable is considered legitimate and is acceptable. For instance, if all the discrepancy between male and female job hiring rates is explained by their education levels, the discrimination can be deemed legitimate and acceptable. 

\textit{\textbf{Emphasis on precision vs recall}}: Precision (the complement of target population error~\citep{dieterich2016compas}) is defined as the fraction of positive instances among the predicted positive instances. In other words, if the system predicts an instance as positive, how precise that prediction is. Recall (the complement of model error~\citep{dieterich2016compas}) is defined as the fraction of the total number of positive instances that are correctly predicted positive. In other words, how many of the positive instances the system is able to identify. There is always a trade-off between precision and recall (increasing one will lead, very often, to decreasing the other). Depending on the scenario at hand, the fairness of the MLDM may be more sensitive to one on the expense of the other. For example, granting loans to the maximum number of deserving applicants contributes more to fairness than making sure that an applicant who has been granted a loan really deserves it\footnote{It is important to mention here that from the loan granting organization's point of view, the opposite is true. That is, it is more important to make sure that an applicant who has been granted a loan really deserves it and will not default in payments because the interest payments resulting from a loan are relatively small compared to the loan amount that could be lost. Our aim here is fairness, while the loan granting organization's goal is benefit.}. When firing employees, however, the opposite is true: fairness is more sensitive to wrongly firing an employee, rather than, firing the maximum number of under-performing employees.

\textit{\textbf{Emphasis on false positive vs false negative}}: Fairness can be more sensitive to false positive misclassification (type I error) rather than false negative  misclassification (type II error), or the opposite. For example, in criminal risk assessment scenario, it is commonly accepted that incarcerating an innocent person (false positive) is more serious than letting a guilty person escape (false negative).

\textit{\textbf{Cost of misclassification}}: Depending on the scenario at hand, the cost of misclassification can be significant (e.g. incarcerating an individual, firing an employee, rejecting a college application, etc.) or mild and without consequential impact (e.g. useless product recommendation, misleading income prediction, offensive online translation, abusive results in online autocomplete, etc.)

\textit{\textbf{Prediction threshold is fixed or floating}}: Decisions in MLDM are typically made based on predicted real-valued score. In the case of binary outcome, the score is turned into a binary value such as $\{0,1\}$ by thresholding\footnote{The threshold is defined by the decision makers depending on the context of interest.}. 
In some scenarios, it is desirable to interpret the real-value score as probability of being accepted (predicted positive). The threshold used as a cutoff point where positive decisions are demarcated from negative decisions can be fixed or floating. A fixed threshold is set carefully and tends to be valid for different datasets and use cases. For instance, in recidivism risk assessment, high risk threshold is typically fixed. A floating threshold can be selected and fine-tuned arbitrarily by practitioners to accommodate a changing context. Acceptance score in loan granting scenarios is an example of a floating threshold as it can move up or down depending on the economic context. {\color{black} When the threshold is floating in a given application, assessing fairness should be done using a suitable fairness notion (e.g. calibration) otherwise, the result of the assessment may be misleading for specific threshold values.}

\textit{\textbf{Likelihood of intersectionality}}: Intersectionality theory~\citep{crenshaw1990mapping} focuses on a specific type of bias due to the combination of sensitive factors. An individual might not be discriminated based on race only or based on gender only, but she might be discriminated because of a combination of both. Black women are particularly prone to this type of discrimination.

\textit{\textbf{Likelihood of masking}}: Masking is a form of intentional discrimination that allows decision makers with prejudicial views to mask their intentions~\citep{barocas2016big}. Masking is typically achieved by exploiting how fairness notions are defined. For example, if the fairness notion requires equal number of candidates to be accepted from two ethnic groups, the MLDM can be designed to carefully select candidates from the first group (satisfying strict requirements) while selecting randomly from the second group just to ``make the numbers''.  

\textcolor{black}{
\textit{\textbf{Sources of Bias}}: Bias in the MLDM outcome can arise from several possible sources at any stage in the data generation and machine learning pipeline. Framing sources of bias necessitates deep understanding of the application at hand and, typically, can only be identified after a "post-mortem" analysis of the predicted outcome. However, in some real-world scenarios, one or more sources of bias may be more likely than others. In such cases, the suspected source of bias can be used as a criterion to select the most appropriate notion for fairness assessment. Sources of bias can be grouped broadly into six categories: historical, representation, measurement, aggregation, evaluation, and deployment~\cite{suresh2019framework}. Historical bias arises when the data reliably collected from the world leads to outcomes which are unwanted and socially unfavorable. For example, while data reliable collected indicates that only 5\% of Fortune 500 CEOs are women~\cite{zarya18}, the resulting outcome of a prediction system based on this data is typically not wanted\footnote{For this reason, Google has changed their image search result for CEO to return a higher proportion of women.}. Representation bias arises when some non-protected populations are under-represented in the training data. Measurement bias arises when the features or label values are not measured accurately. For example, Street Bump is an application used in Boston city to detect when residents drive over potholes thanks to the accelerometers built into smartphones~\cite{crawford2013think}. Collecting data using this application introduces a measurement bias due to the disparity in the distribution of smartphones according to the different districts in the city, which are often correlated with race or level of income. Aggregation bias arises when sub-populations are aggregated together while a single model is unlikely to fit all sub-populations. For instance, the genetic risk scores derived largely on European populations have been shown to generally perform very poorly in the prediction of osteoporotic fracture and bone mineral density on non-European populations, in particular, on Chinese population~\cite{li2019genetic}. Evaluation bias arises when the training data differs significantly from the testing data. For instance, several MLDMs are trained using benchmark datasets which may be very different from the target dataset. Deployment bias arises when there is a disparity between the initial purpose of an MLDM and the way it is actually used. For instance, a child maltreatment MLDM might be designed to predict the risk of child abuse after two years from the reception of a referral call, while in practice it may be used to help social agents take decisions about an intervention. This can lead to a bias since the decision has an impact on the outcome~\cite{coston2020counterfactual}.
}

\textcolor{black}{
\textit{\textbf{Legal Framework}}:
Anti-discrimination regulations in several countries, in particular US, distinguish between two legal frameworks, namely disparate treatment and disparate impact~\cite{barocas2016big}. In the disparate treatment framework, a decision is considered unfair if it uses (directly or indirectly) the individual's sensitive attribute information. In the disparate impact framework, a decision is unfair if it results in an outcome that is disproportionately disadvantageous (or beneficial) to individuals according to their sensitive attribute information. Zafar et al.~\cite{zafar2017fairness} formalized another fairness criterion, namely, disparate mistreatment according to which, a decision is unfair if it results in different misclassification rates for groups of people with different sensitive attribute information. Note that this criterion is currently not supported by a legal framework. Machine learning fairness notions can be classified according to the type of fairness it is evaluating. For instance, if a plaintiff is accusing an employer for intentional discrimination, she should consider the disparate treatment legal framework, and hence a fairness notion which falls in that framework.
}

\textit{\textbf{The existence of regulations and standards}}: In some domains, laws and regulations might be imposed to avoid discrimination and bias. 
For instance, guidelines from the \textit{U.S. Equal Employment Opportunity Commission} state that a difference of the probability of acceptance between two sub-populations exceeding $20\%$ is illegal \citep{barocas-hardt-narayanan}. Another example might be an internal organizational policy imposing diversity among its employees.

\section {Fairness notions}
\label{sec:notions}

Let $V$, $A$, and $X$ be three random variables representing, respectively, the total set of attributes, the sensitive attributes, and the remaining attributes describing an individual such that $V=(X,A)$ and $P(V=v_i)$ represents the probability of drawing an individual with a vector of values $v_i$ from the population. For simplicity, we focus on the case where $A$ is a binary random variable where $A=0$ designates the protected group, while $A=1$ designates the non-protected group. Let $Y$ represent the actual outcome and $\hat{Y}$ represent the outcome returned by the prediction algorithm (MLDM). Without loss of generality, assume that $Y$ and $\hat{Y}$ are binary random variables where $Y=1$ designates a positive instance, while $Y=0$ a negative one. A perfect MLDM will match perfectly the actual outcome ($\hat{Y} = Y$). Typically, the predicted outcome $\hat{Y}$ is derived from a score represented by a random variable $S$ where $P(S = s)$ is the probability that the score value is equal to $s$.

All fairness notions presented in this section address the following question: ``is the outcome/prediction of the MLDM fair towards individuals?''. So fairness notion is defined as a mathematical condition that must involve either $\hat{Y}$ or $S$ along with the other random variables. As such, we are not concerned by the inner-workings of the MLDM and their fairness implications. What matters is only the score/prediction value and how fair/biased it is.

Most of the proposed fairness notions are properties of the joint distribution of the above random variables ($X$, $A$, $Y$, $\hat{Y}$, and $S$). They can also be interpreted using the confusion matrix and the related metrics (Table~\ref{tab:confMat}).

\begin{table}[!ht]
\caption{Metrics based on confusion matrix. \medskip}
\label{tab:confMat}      
\centering
\begin{tabular}{|L|l|l|l|l|}
\hline
{\smallskip}{\smallskip}{\smallskip}{\smallskip}
 & \makecell{Actual Positive}    & \makecell{Actual Negative} & & \\
{\smallskip}{\smallskip}{\smallskip}{\smallskip}
 & \makecell{$Y = 1$}  & \makecell{$Y = 0$} & & \\
\hline
   \makecell{Predicted\\ Positive}  & \makecell{\textbf{TP}\\ {\small(True Positive)}}& \makecell{ \textbf{FP} \\{\small (False Positive)}} &\textbf{PPV} =  $\frac{TP}{TP+FP}$ & \textbf{FDR} =  $\frac{FP}{TP+FP}$ \\
    $\hat{Y} = 1$ & & {\small\textit{Type I error}}  &{\small\textit{Positive Predictive Value} }& {\small\textit{False Discovery Rate}}\\
     & &  &{\small\textit{Precision}} & {\small\textit{Target Population Error}} \\
          & &  & {\small\textit{PV+}} & \\
             & &  & {\small\textit{Target Population Error}}&\\
   \hline
  \makecell{Predicted\\ Negative} & \makecell{ \textbf{FN}\\ (False Negative)}& \makecell{\textbf{TN} \\(True Negative)}&\textbf{FOR} =  $\frac{FN}{FN+TN}$  &\textbf{NPV} =  $\frac{TN}{FN+TN}$\\ 
$\hat{Y} = 0$&{\small\textit{Type II error}} & &  {\small\textit{False Omission Rate} } & {\small\textit{Negative Predictive Value}} \\
&& &{\small\textit{Success Predictive Error}}  &  {\small\textit{PV-}}\\

  \hline
    \rule{0pt}{4ex}              \makecell{\\ \textcolor{white}{ww} }   &\textbf{TPR} = $\frac {TP}{TP+FN}$ &\textbf{FPR} = $\frac{FP}{FP+TN}$ &\textbf{OA} = $\frac{TP+TN}{TP+FP+TN+FN}$& \textbf{BR} = $\frac{TP+FN}{TP+FP+TN+FN}$\\
     &{\small\textit{True Positive Rate} }& {\small\textit{False Positive Rate} }  & {\small\textit {Overall Accuracy}} & {\small \textit{Base Rate}}\\
                & {\small\textit{Sensitivity}} & {\small\textit{Model Error}}  & &{\small\textit{Prevalence ($p$)}} \\
                & {\small\textit{Recall}} &  & & \\
                            \hline
            \rule{0pt}{4ex}    \makecell{\\ \textcolor{white}{ww} }       &\textbf{FNR} = $\frac{FN}{TP+FN}$ & \makecell { \textbf{TNR} = $\frac{TN}{FP+TN}$} & &  \\
                & {\small\textit {False Negative Rate} }&{\small\textit{True Negative Rate}}  & & \\
                & {\small\textit{Model Error}}& {\small\textit{Specificity} }& & \\
 \hline
\end{tabular}
\end{table}

While presenting and discussing fairness notions, whenever needed, we use the simple job hiring scenario of Table~\ref{tab:example1}. Each sample in the dataset has the following attributes: education level (numerical), job experience (numerical), age (numerical), marital status (categorical), gender (binary) and a label (binary). The sensitive attribute is the applicant gender, that is, we are focusing on whether male and female applicants are treated equally. Table~\ref{tab:example1}(b) presents the predicted decision  (first column) and the predicted score value (second column) for each sample. The threshold value is set to $0.5$.

\begin{table}[!ht]
	\centering
	\caption{A simple job hiring example. $Y$ represents the data label indicating whether the applicant is hired ($1$) or rejected ($0$). $\hat{Y}$ is the prediction which is based on the score $S$. A threshold of $0.5$ is used. \medskip}
\label{tab:example1} 
\subfloat[Dataset]{
 \begin{tabular}{ @{} cLMcLc@{} }
\hline\noalign{\smallskip}
    Gender    & Education Level & Job Experience &  Age  & Marital Status & Y \\
    \noalign{\smallskip}\hline\noalign{\smallskip}
    Female 1           & 8    & 2 &39            & single    & 0 \\
   Female 2            & 8     & 2   & 26             & married     & 1\\
    Female 3            & 12     & 8   & 32             & married     & 1\\
    Female 4     & 11      &  3   & 35             & single     & 0  \\
    Female 5     & 9      &  5   & 29             & married     & 1 \\
    Male 1          & 11    & 3 &34            & single    & 1  \\
    Male 2          & 8     & 0   & 48             & married     & 0 \\
    Male 3        & 7     & 3   & 43             & single     & 1   \\
    Male 4      & 8      &  2   & 26             & married     & 1 \\
    Male 5          &8     & 2   & 41             & single     & 0  \\
    Male 6   & 12      &  8   & 30             & single     & 1  \\
    Male 7   & 10      &  2   & 28             & married     & 1 \\
\noalign{\smallskip}\hline
  \end{tabular}  
  }
 \qquad 
  \subfloat[Prediction]{
   \begin{tabular}{ @{} NN@{} }
\hline\noalign{\smallskip}
    \^{Y} & S\  \\
    \\
\noalign{\smallskip}\hline\noalign{\smallskip}
    1 & 0.5\\
    0 & 0.1\\
    1 & 0.5\\
    0 & 0.2\\
    0 & 0.3\\
    1 & 0.8\\
    0 & 0.1\\
    0 & 0.1\\
    1 & 0.5\\
    1 & 0.5\\
    1 & 0.8\\
    0 & 0.3\\
\noalign{\smallskip}\hline
  \end{tabular}
}   
\end{table}

A simple and straightforward approach to address fairness problem is to ignore completely any sensitive attribute while training the MLDM system. This is called \textit{fairness through unawareness}\footnote{Known also as: blindness, unawareness \citep{mitchell2018prediction}, anti-classification \citep{corbett2018measure}, and treatment parity \citep{lipton2018does}.}. We don't treat this approach as fairness notion since, given MLDM prediction, it does not allow to tell if the MLDM is fair or not. Besides, it suffers from the basic problem of proxies. Many attributes (e.g. home address, neighborhood, attended college) might be highly correlated to the sensitive attributes (e.g. race) and act as proxies of these attributes. Consequently, in almost all situations, removing the sensitive attribute during the training process does not address the problem of fairness.

\subsection{Statistical parity}
\label{sec:notion_sp}
Statistical parity \citep{dwork2012fairness} (a.k.a demographic parity \citep{kusner2017counterfactual}, independence \citep{barocas2017fairness}, equal acceptance rate \citep{zliobaite2015relation}, benchmarking \citep{simoiu2017problem}, group fairness \citep{dwork2012fairness}) is one of the most commonly accepted notions of fairness. It requires the prediction to be statistically independent of  the sensitive attribute $(\hat{Y}  \perp A)$. Thus, a classifier \^{Y} satisfies statistical parity if:
\begin{equation}
\label{eq:sp}
P(\hat{Y} \mid A = 0) = P(\hat{Y} \mid A = 1) 
\end{equation}

In other words, the predicted acceptance rates for both protected and unprotected groups should be equal. Using the confusion matrix (Table~\ref{tab:confMat}), statistical parity implies  that {\small $(TP+FP)/(TP+FP+FN+TN)$} should be equal for both groups. In the MLDM of Table~\ref{tab:example1}, it means that one should not hire proportionally more applicants from one group than the other. The calculated predicted acceptance rate of hiring male and female applicants is $0.57$ ($4$ out of $7$) and $0.4$ ($2$ out of $5$), respectively. Thus, the MLDM of Table~\ref{tab:example1} does not satisfy statistical parity.

Statistical parity is appealing in scenarios where there is a preferred decision over the other,{ \color{black}and provided there are no other considerations relevant for the decision, in which case, the following fairness notion namely, conditional statistical parity, is more suitable}. For example, being accepted to a job, not being arrested, being admitted to a college, etc.\footnote{This might not be the case in other scenarios such as disease prediction, child maltreatment, where imposing a parity of positive predictions is meaningless.}. What really matters is a balance in the prediction rate among all groups.  

Statistical parity is suitable when the label $Y$ is not trustworthy due to some flawed or biased measurement\footnote{This is also known as differential measurement error \citep{vanderweele2012results}.}. An example of this type of problem was observed in the recidivism risk
prediction tool COMPAS \citep{angwin2016machine}. Because minority groups are more controlled, and more officers are dispatched in their regions, the number of arrests (used to assess the level of crime~\citep{suresh2019framework}) of those minority groups is significantly higher than that of the rest of the population. Hence, for fairness purposes, in the absence of information to precisely quantify the differences in recidivism by race, the most suitable approach is to treat all sub-populations equally with respect to recidivism \citep{johndrow2019algorithm}.

Statistical parity is also well adapted to contexts in which some regulations or standards are imposed. For example, a law might impose to equally hire or admit applicants from different sub-populations.  

\begin{figure}[!ht]
\centering
\includegraphics [scale=0.4] {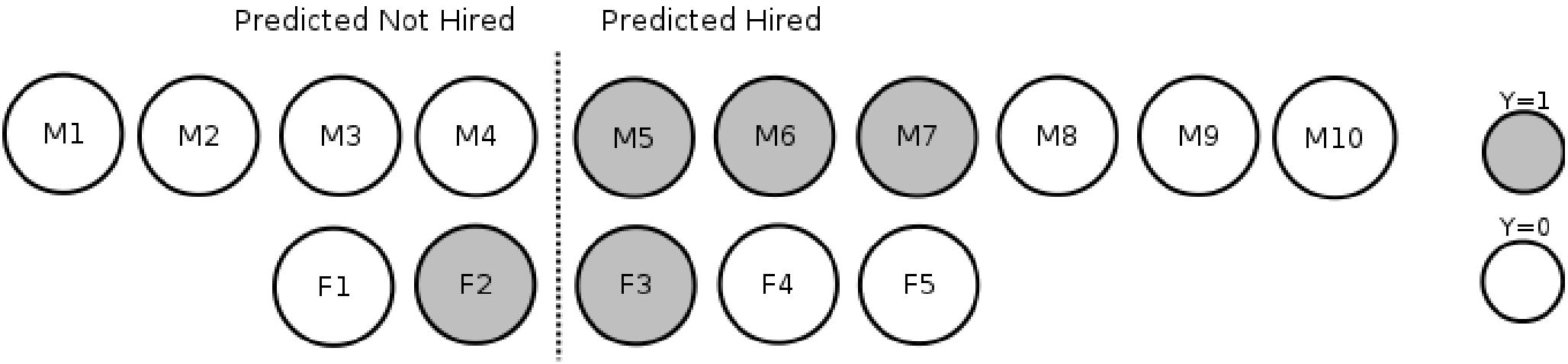}
\caption{$F_i$ and $M_i$ ($i \in [1-10]$) designate female and male applicants, respectively. The grey shaded circles indicate applicants who belong to the positive class while white circles indicate applicants belonging to the negative class. The dotted vertical line is the prediction boundary. Thus, applicants at the right of this line are predicted hired while applicants at the left are predicted not hired.}
\label{fig:spExample}     
\end{figure}

The main problem of statistical parity is that it doesn't consider a potential correlation between the label $Y$ and the sensitive attribute $A$. In other words, if the underlying base rates of the protected and unprotected groups are different, statistical parity will be misleading. In particular, modifying an MLDM with a perfect prediction ($\hat{y}=y$) so to satisfy statistical parity while the base rates are different will lead to loss of utility~\citep{hardt2016equalityshort}. As an example, Figure~\ref{fig:spExample} illustrates a scenario for hiring computer engineers where equal proportions of male/female applicants have been predicted hired ($60\%$) thus, satisfying statistical parity. However, when considering the label and more precisely the base rates that differ in both groups ($0.3$ for men versus $0.4$ for women), the classifier becomes discriminative against female applicants ($50\%$ of qualified female applicants are not predicted hired). More generally, when the ground truth is available and is used in the training of the MLDM, statistical parity is not recommended because, very often, it conflicts with the ground truth
~\citep{zafar2017fairness}. 

Another issue with this notion is its ``laziness''; if we hire carefully selected applicants from male group and random applicants from female group, we can still achieve statistical parity, yet leading to negative results for the female group as its performance will tend to be worse than that of male group. This practice is an example of \textit{self-fulfilling prophecy}~\citep{dwork2012fairness} where a decision maker may simply select random members of a protected group rather than qualified ones, and hence, intentionally building a bad track record for that group. Barocas and Selbst refer to this problem as masking~\citep{barocas2016big}. Masking is possible to game several fairness notions, but it is particularly easy to carry out in the case of statistical parity.

\subsection{Conditional statistical parity}
\label{sec:notioncsp}
Conditional statistical parity~\citep{corbett2017algorithmic}, called also conditional discrimination-aware classification in~\citep{kamiran2013quantifying} is a variant of statistical parity obtained by controlling on a set of legitimate attributes\footnote{Called explanatory attributes in \citep{kamiran2013quantifying}.}. The legitimate attributes (we refer to them as $E$) among $X$ are correlated with the sensitive attribute $A$ and give some factual information about the label at the same time leading to a \textit{legitimate} discrimination. In other words, this notion removes the illegal discrimination, allowing the disparity in decisions to be present as long as they are explainable~\citep{corbett2017algorithmic}. In the hiring example, possible explanatory factors that might affect the hiring decision for an applicant  could be the education level and/or the job experience. If the data is composed of many highly educated and experienced male applicants and only few highly educated and experienced women,  one might justify the disparity between predicted acceptance rates between both groups and consequently, does not necessarily reflect gender discrimination. 
Conditional statistical parity holds if:
\begin{equation}
\label{eq:csp}
P(\hat{Y}=1 \mid E=e,A = 0) = P(\hat{Y}=1 \mid E=e,A = 1) \quad \forall e
\end{equation}

\begin{table}[!ht]
\centering
\caption{Application of conditional statistical parity by controlling on education level and job experience. \medskip}
\label{tab:condSP} 
\subfloat[Dataset]{
 \begin{tabular}{ @{} cLMcLc@{} }
\hline\noalign{\smallskip}
    Gender    & Education Level & Job Experience &  Age  & Marital Status & $Y$ \\
    \noalign{\smallskip}\hline\noalign{\smallskip}
    Female 1           & 8    & 2 &39            & single    & 0 \\
   Female 2            & 8     & 2   & 26      & married     & 1\\
   Female 3            & 12     & 8   & 32      & married     & 1\\
   Male 4      & 8      &  2   & 26             & married     & 1 \\
    Male 5          &8     & 2   & 41             & single     & 0  \\
    Male 6          &12     & 8   & 30             & single     & 1  \\
   \noalign{\smallskip}\hline
  \end{tabular}  
  }
 \qquad 
  \subfloat[Prediction]{
   \begin{tabular}{ @{} NN@{} }
\hline\noalign{\smallskip}
    \^{Y} & S\  \\
    \\
\noalign{\smallskip}\hline\noalign{\smallskip}
    1 & 0.5\\
    0 & 0.1\\
    1 & 0.5\\
    1 & 0.5\\
     1 & 0.5\\
    1 & 0.8\\
\noalign{\smallskip}\hline
  \end{tabular}
}   
\end{table}

Table~\ref{tab:condSP} shows two possible combinations values for $E$. The first combination (education level=$8$ and job experience=$2$) includes samples Female~1, Female~2, Male~4, and Male~5 for which the prediction is clearly discriminative against women as the predicted acceptance rates for men and women are $1$ and $0.5$, respectively. The second combination (education level=$12$ and job experience=$8$) includes Female~3 and Male~6 in which the prediction is fair (predicted acceptance rate is $1$ for both applicants). Overall, the prediction is not fair as it does not hold for one combination of values of $E$.

In practice, conditional statistical parity is suitable when there is one or several attributes that justify a possible disparate treatment between different groups in the population. Hence, choosing the legitimate attribute(s) is a very sensitive issue as it has a direct impact on the fairness of the decision-making process. More seriously, conditional statistical parity gives a decision maker a tool to game the system and realize a self-fullfilling prophecy. Therefore, it is recommended to resort to domain experts or law officers to decide what is unfair and what is tolerable to use as legitimate discrimination attribute~\citep{kamiran2013quantifying}.

\subsection{Equalized odds}
\label{sec:eqOdds}
Unlike the two previous notions, equalized odds~\citep{hardt2016equality} (separation in~\citep{barocas2017fairness}, conditional procedure accuracy equality in~\citep{berk2018fairness}, disparate mistreatment in~\citep{zafar2017fairness}, error rate balance in~\citep{chouldechova2017fair}) considers both the predicted and the actual outcomes. Thus, the prediction is conditionally independent from the protected attribute, given the actual outcome $(\hat{Y} \perp A \mid Y)$.
In other words, equalized odds requires both sub-populations to have the same TPR and FPR~(Table~\ref{tab:confMat}). In our example, this means that the probability of an applicant who is actually hired to be predicted hired and the probability of an applicant who is actually not hired to be incorrectly predicted hired should be both same for men and women: 

\begin{equation}
\label{eq:eqOdds}
P(\hat{Y} = 1 \mid Y=y,\; A=0) = P(\hat{Y}=1 \mid Y= y,\; A=1)  \quad \forall{ y \in \{0,1\}}
\end{equation}

In the example of Table~\ref{tab:example1}, the TPR for male and female groups is $0.6$ and $0.33$, respectively while the FPR is exactly the same ($0.5$) for both groups. Consequently, the equalized odds does not hold.

By contrast to statistical parity, equalized odds is well-suited for scenarios where the ground truth exists such as: disease prediction or stop-and-frisk~\citep{bellin2014inverse}. It is also suitable when the emphasis is on recall (the fraction of the total number of positive instances that are correctly predicted positive) rather than precision (making sure that a predicted positive instance is actually a positive instance). 

A potential problem of equalized odds is that it may not help closing the gap between the protected and unprotected groups. For example, consider a group of $20$ male applicants of which $16$ are qualified and another equal size group of $20$ females of which only $2$ are qualified. If the employer decides to hire $9$ applicants and while satisfying equalized odds, $8$ offers will be granted to the male group and only $1$ offer will be granted to the female group. While this decision scheme looks fair on the short term, on the long term, however, it will contribute to confirm this ``unfair'' status-quo and perpetuate this vicious cycle\footnote{If the job is a well-paid, male group tends to have a better living condition and affords better education for their kids, and thus enable them to be qualified for such well-paid jobs when they grow up. The gap between the groups will tend to increase over time.}. Whether to consider this long term impact as a problem of equalized odds is a controversial issue as it overlaps with the different but related question of ``how to address unfairness?''. Note that other fairness notions, such as statistical parity, help closing the gap between the protected and unprotected groups on the long term.

Because equalized odds requirement is rarely satisfied in practice, two variants can be obtained by relaxing Eq.~\ref{eq:eqOdds}. The first one is called \textbf{equal opportunity}~\citep{hardt2016equality} (false negative error rate balance in~\citep{chouldechova2017fair}) and is obtained by requiring only TPR equality among groups: 
\begin{equation}
\label{eq:eqOpp}
P(\hat{Y}=1 \mid Y=1,A = 0) = P(\hat{Y}=1\mid Y=1,A = 1) 
\end{equation}

In the job hiring example, this is to say that we should hire equal proportion of individuals from the qualified fraction of each group.

As $TPR = TP/(TP+FN)$ (Table~\ref{tab:confMat}) does not take into consideration $FP$, equal opportunity is completely insensitive to the number 
of false positives. This is an important criterion when considering this fairness notion in practice. More precisely, in scenarios where a disproportionate number of false positives among groups has fairness implications, equal opportunity should not be considered. The scenario in Table~\ref{tab:eqOpp} shows an extreme case of a job hiring dataset where the male group has a large number of false positives (Male $7 - 100$) while equal opportunity is satisfied. 

\begin{table}[!ht]
\centering
\caption{An extreme job hiring scenario satisfying equal opportunity. All Male $7 - 100$ samples are false positives (label $Y$ is $0$ and prediction $\hat{Y}$ is $1$). \medskip}
\label{tab:eqOpp} 
\subfloat[Dataset]{
 \begin{tabular}{ @{} cLMcLc@{} }
\hline\noalign{\smallskip}
    Gender    & Education Level & Job Experience &  Age  & Marital Status & $Y$ \\
    \noalign{\smallskip}\hline\noalign{\smallskip}
    Female 1           & 8    & 2 &39            & single    & 1 \\
   Female 2            & 8     & 2   & 26      & married     & 0\\
   Female 3            & 12     & 8   & 32      & married     & 1\\
   Male 4      & 8      &  2   & 26             & married     & 1 \\
    Male 5          &8     & 2   & 41             & single     & 0  \\
    Male 6          &12     & 8   & 30             & single     & 1  \\
    Male 7          &10     & 5   & 32             & married     & 0  \\
    $\ldots$          & $\ldots$     & $\ldots$   & $\ldots$     & $\ldots$     & 0  \\
    Male 100          &8     & 10   & 27             & single     & 0  \\
   \noalign{\smallskip}\hline
  \end{tabular}  
  }
 \qquad 
  \subfloat[Prediction]{
   \begin{tabular}{ @{} NN@{} }
\hline\noalign{\smallskip}
    \^{Y} & S\  \\
    \\
\noalign{\smallskip}\hline\noalign{\smallskip}
    1 & 0.5\\
    0 & 0.1\\
    0 & 0.3\\
    1 & 0.5\\
     0 & 0.2\\
    0 & 0.4\\
    1 & 0.8\\
    1 & $\ldots$\\
    1 & 0.7\\
\noalign{\smallskip}\hline
  \end{tabular}
}   
\end{table}

To decide about the suitability of equal opportunity in the job hiring example, the question that should be answered by stakeholders and decision makers is ``if all other things are equal, is it fair to hire disproportionately more unqualified male candidates?''. For the employer, it is undesirable to have several false positives (regardless of their gender) as the company will end up with unqualified employees. For a stakeholder whose goal is to guarantee fairness between males and females, it is not very critical to have more false positives in one group, provided that these two groups have the same proportion of false negatives (a qualified candidate which is not hired). 

In the scenario of predicting which employees to fire, however, a false positive (firing a well-performing employee) is critical for fairness. Hence, equal opportunity should not be used as a measure of fairness.

The second relaxed variant of equalized odds is called \textbf{predictive equality}~\citep{corbett2017algorithmic} (false positive error rate balance in~\citep{chouldechova2017fair}) which requires only the FPR to be equal in both groups. 

In other words, predictive equality checks whether the accuracy of decisions is equal across protected and unprotected groups:
\begin{equation}
\label{eq:predEq}
P(\hat{Y}=1 \mid Y=0,A = 0) = P(\hat{Y}=1\mid Y=0,A = 1) 
\end{equation}
In the job hiring example, predictive equality holds when the probability of an applicant with an actual weak profile for the job to be incorrectly predicted hired is the same for both men and women.

Since $FPR = FP/(FP+TN)$ (Table~\ref{tab:confMat}) is independent from $FN$, predictive equality is completely insensitive to false negatives. One can come up with an extreme example similar to Table~\ref{tab:eqOpp} with a disproportionate number of false negatives but predictive equality will still be satisfied (keeping all other rates equal). Hence, in scenarios where fairness between groups is sensitive to false negatives, predictive equality should not be used. Such scenarios include hiring and admission where a false negative means a qualified candidates are rejected disproportionately among groups. Predictive equality is acceptable in criminal risk assessment scenarios as false negatives (releasing a guilty person) are less critical than false positives (incarcerating an innocent person).   

Predictive equality is particularly suitable to measure the fairness of face recognition systems in crime investigation where security camera footage are analyzed. Fairness between ethnic groups with distinctive face features is very sensitive to the FPR. A false positive means an innocent person is being flagged as participating in a crime. If this false identification happens at a much higher rate for a specific sub-population (e.g. dark skinned group) compared to the rest of the population, it is clearly unfair for individuals belonging to that sub-population.

Looking to the problem from another perspective, choosing between equal opportunity and predictive equality depends on how the outcome/label is defined. In scenarios where the positive outcome is desirable (e.g. hiring, admission), typically fairness is more sensitive to false negatives rather than false positives, and hence equal opportunity is more suitable. In scenarios where the positive outcome is undesirable for the subjects (e.g. firing, risk assessment), typically fairness is more sensitive to false positives rather than false negatives, and hence predictive equality is more suitable.

The following proposition states formally the relationship between equalized odds, equal opportunity, and predictive equality.

\begin{proposition}
Satisfying equal opportunity and predictive equality is equivalent to satisfying equalized odds:
$$Eq.~\ref{eq:eqOdds} \Leftrightarrow Eq.~\ref{eq:eqOpp} \wedge Eq.~\ref{eq:predEq}$$
\end{proposition}

\subsection{Conditional use accuracy equality}
\label{sec:condUseAcc}
Conditional use accuracy equality \citep{berk2018fairness} (called sufficiency in~\citep{barocas2017fairness}) is achieved when all population groups have equal $PPV=\frac{TP}{TP+FP}$ and $NPV=\frac{TN}{FN+TN}$.  In other words, the probability of subjects with positive predictive value to truly belong to the positive class and the probability of subjects with negative predictive value to truly belong to the negative class should be the same:

\begin{equation} 
\label{eq:condUseAcc}
P(Y=y\mid \hat{Y}=y ,A = 0) = P(Y=y\mid \hat{Y}=y,A = 1) \quad \forall{ y \in \{0,1\}}
\end{equation}

Intuitively, this definition implies equivalent accuracy for male and female applicants from both positive and negative predicted classes \citep{verma2018fairness}. By contrast to equalized odds (Section~\ref{sec:eqOdds}), one is conditioning on the algorithm’s predicted outcome not the actual outcome.  In other words, this notion emphasis the precision of the MLDM system rather than its sensitivity (a trade-off discussed earlier in Section~\ref{sec:criteria}).

The calculated PPVs for male and female applicants in our hiring example (Table~\ref{tab:example1}) are $0.75$ and $0.5$, respectively. NPVs for male and female applicants are both equal to $0.33$. Overall the dataset in Table~\ref{tab:example1} does not satisfy conditional use accuracy equality. 

\textbf{Predictive parity}~\citep{chouldechova2017fair} (called outcome test in ~\citep{simoiu2017problem}) is a relaxation of conditional use accuracy equality requiring only equal PPV among groups:

\begin{equation}
\label{eq:predPar}
P(Y=1 \mid \hat{Y} =1,A = 0) = P(Y=1\mid \hat{Y} =1,A = 1)  
\end{equation}

In our example, this is to say that the prediction used to determine the candidate’s eligibility for a particular job should reflect the candidate’s actual capability of doing this job which is harmonious with the employer’s benefit. 

Like predictive equality (Eq.~\ref{eq:predEq}), predictive parity is insensitive to false negatives. Hence in any scenario where fairness is sensitive to false negatives, predictive parity should not be considered sufficient. 

Choosing between predictive parity and equal opportunity depends on whether the scenario at hand is more sensitive to precision or recall. For precision-sensitive scenarios, typically predictive parity is more suitable while for recall-sensitive scenarios, equal opportunity is more suitable. Precision-sensitive scenarios include disease prediction, child maltreatment risk assessment, and firing from jobs. Recall-sensitive scenarios include loan granting, recommendation systems, and hiring. Very often, precision-sensitive scenarios coincide with situations where the positive prediction ($\hat{Y}=1$) entails a higher cost~\citep{zafar2017fairness}. For example, a predicted child maltreatment case will result in placing the child in a foster house which will generally entail a higher cost compared to a negative prediction (low risk of child maltreatment) in which case the child stays with the family and typically no action is taken.  

Conditional use accuracy equality (Eq.~\ref{eq:condUseAcc}) is ``symmetric'' to equalized odds (Eq.~\ref{eq:eqOdds}) with the only difference of switching $Y$ and $\hat{Y}$. The same holds for equal opportunity (Eq.~\ref{eq:eqOpp}) and predictive parity (Eq.~\ref{eq:predPar}). However, there is no ``symmetric'' notion to predictive equality (Eq.~\ref{eq:predEq}). For completeness, we define such notion and give it the name  \textbf{negative predictive parity}.

\begin{definition}
Negative predictive parity holds iff all sub-groups have the same $NPV = \frac{TN}{FN+TN}$:
\begin{equation}
\label{eq:negpredEq}
P(Y=1 \mid \hat{Y} =0,A = 0) = P(Y=1\mid \hat{Y} =0,A = 1)  
\end{equation}
\end{definition}

The following proposition states formally the relationship between conditional use accuracy equality, predictive parity, and negative predictive parity.

\begin{proposition}
Satisfying predictive parity and negative predictive parity is equivalent to satisfying conditional use accuracy equality:
$$Eq.~\ref{eq:condUseAcc} \Leftrightarrow Eq.~\ref{eq:predPar} \wedge Eq.~\ref{eq:negpredEq}$$
\end{proposition}

\subsection{Overall accuracy equality}
\label{sec:ovAcc}
Overall accuracy equality ~\citep{berk2018fairness} is achieved when overall accuracy for both groups is the same. Thus, true negatives and true positives are equally considered and desired.  Using the confusion matrix (Table~\ref{tab:confMat}), this implies that {\small $(TP+TN)/(TP+FN+FP+TN)$} is equal for both groups. In our example, it is to say that the probability of well-qualified applicants to be correctly accepted for the job and non-qualified applicants to be correctly rejected is the same  for both male and female applicants:
\begin{equation}
\label{eq:ovAcc}
P(\hat{Y} = Y | A = 0) = P(\hat{Y} = Y | A = 1) 
\end{equation}

\begin{table}[!h]
	\centering
	\caption{A job hiring scenario satisfying overall accuracy but not conditional use accuracy equality. \medskip}
\label{tab:oa} 
\subfloat{
 \begin{tabular}{ @{} rcl@{} }
&&\\
&&\\
&&\\
&&\\
OA&=& 0.625\\
PPV&=& 1 \\
NPV&=& 0.25 \\
&&\\
&&\\
  \end{tabular} 
  }
\captionsetup[subfloat]{position=top,labelformat=empty}
\subfloat[Group 1 (Female)]{
 \begin{tabular}{ @{} ccc@{} }
\hline\noalign{\smallskip}
Gender & $\; Y \; $    & $\; \hat{Y} \; $ \\
    \noalign{\smallskip}\hline\noalign{\smallskip}
    F1 & 1 & 1\\
    F2 & 1 & 0\\
    F3 & 1 & 0\\
    F4 & 0 & 0\\
    F5 & 1 & 1\\
    F6 & 1 & 1\\
    F7 & 1 & 0\\
    F8 & 1 & 1\\
   \noalign{\smallskip}\hline
  \end{tabular}  
  }
 \qquad 
 \subfloat[Group 2 (Male)]{
   \begin{tabular}{ @{} ccc@{} }
\hline\noalign{\smallskip}
Gender & $\; Y \; $ & $\; \hat{Y} \; $  \\
\noalign{\smallskip}\hline\noalign{\smallskip}
    M1 & 1 & 1\\
    M2 & 0 & 1\\
    M3 & 0 & 1\\
    M4 & 0 & 0\\
    M5 & 0 & 0\\
    M6 & 0 & 0\\
    M7 & 0 & 1\\
    M8 & 1 & 1\\
\noalign{\smallskip}\hline
  \end{tabular}
}   
\subfloat{
 \begin{tabular}{ @{} rcl@{} }
 &    & \\
     &  & \\
     &  & \\
     &  & \\
     OA & =  & 0.625\\
     PPV & = & 0.4 \\
     NPV & =  & 1 \\
     &  & \\
     &  & \\
  \end{tabular} 
  }

\end{table}

Overall accuracy equality is closely related to equalized odds (Eq.~\ref{eq:eqOdds}) and to conditional use accuracy equality (Eq.~\ref{eq:condUseAcc}). The main difference is that overall accuracy equality aggregates together positive class and negative class misclassifications (FP and FN). Aggregating together FP and FN (and hence TP and TN) without any distinction is very often misleading for fairness purposes. 

\begin{proposition}
An MLDM that satisfies equalized odds or conditional use accuracy equality always satisfies overall accuracy. 
$$Eq.~\ref{eq:eqOdds} \vee Eq.~\ref{eq:condUseAcc} \Rightarrow Eq.~\ref{eq:ovAcc} $$
\end{proposition}

The reverse, however, is not true. That is, an MLDM that satisfies overall accuracy does not necessarily satisfy equalized odds or conditional use accuracy equality. To prove it, consider the example in Table~\ref{tab:oa} satisfying overall accuracy equality but not conditional use accuracy equality. For the female group, there are only FN misclassifications (no FP) and more TPs than TNs, while in the male group, there are only FP misclassifications (no FN) and more TNs than TPs. But since the proportion of correct classifications is the same in both groups (5 out of 8), overall accuracy equality holds. 
In real-world applications, it is very uncommon that TP (or FN) and TN (or FP) are desired at the same time and without distinction. \textcolor{black}{For example, overall accuracy equality is not suitable to measure fairness in child maltreatment prediction because a False Positive (misclassifying a child case which is not at risk\footnote{Results in a useless intervention, because the child is not at risk anyway.}) is less damaging than a False Negative (misclassifying a child case which is at risk\footnote{Results in a failure to anticipate a child maltreatment.}). A hypothetical health care scenario where overall accuracy equality is suitable is when both types of misclassifications have the same cost/benefit. For example, an eventual health condition that yields very similar complications (1) when the treatment is administered wrongly and (2) when the treatment is not administered while it is needed.}  

\subsection{Treatment equality}
 \label{sec:treatEq}
Treatment equality \citep{berk2018fairness} is achieved when the ratio of FPs and FNs is the same for both protected and unprotected groups:

\begin{equation}
\label{eq:treatEq}
\frac{FN}{FP} \textsubscript{(\textcolor{black}{A=0)}} = \frac {FN}{FP} \textsubscript{(\textcolor{black}{A=1)}}
\end{equation}

Treatment equality is insensitive to the numbers of TPs and TNs which are important to identify bias between sub-populations in most real-world scenarios. Berk et al.~\citep{berk2018fairness} note that treatment equality can serve as an indicator to achieve other kinds of fairness. Table~\ref{tab:te} shows a dataset which fails to satisfy all previous notions, yet, treatment equality is satisfied. Treatment equality can be used in real-world scenarios where only the type of rate of misclassification matters for fairness.

\begin{table}[!h]
	\centering
	\caption{A job hiring scenario satisfying treatment equality but not satisfying all of the previous notions. \medskip}
\label{tab:te} 
\subfloat{
 \begin{tabular}{ @{} rcl@{} }
&&\\
&&\\
&&\\
TPR&=&0.33\\
FPR&=& 0.8\\
PPV&=& 0.2 \\
NPV&=& 0.33 \\
OA&=& 0.25 \\
FN/FP&=& 0.5\\
  \end{tabular} 
  }
  \captionsetup[subfloat]{position=top,labelformat=empty}
\subfloat[Group 1 (Female)]{
 \begin{tabular}{ @{} ccc@{} }
\hline\noalign{\smallskip}
Gender & $\; Y \; $    & $\; \hat{Y} \; $ \\
    \noalign{\smallskip}\hline\noalign{\smallskip}
    F1 & 1 & 1\\
    F2 & 0 & 0\\
    F3 & 0 & 1\\
    F4 & 0 & 1\\
    F5 & 0 & 1\\
    F6 & 0 & 1\\
    F7 & 1 & 0\\
    F8 & 1 & 0\\
   \noalign{\smallskip}\hline
  \end{tabular}  
  }
 \qquad 
 \subfloat[Group 2 (Male)]{
   \begin{tabular}{ @{} ccc@{} }
\hline\noalign{\smallskip}
Gender & $\; Y \; $ & $\; \hat{Y} \; $  \\
\noalign{\smallskip}\hline\noalign{\smallskip}
    M1 & 1 & 1\\
    M2 & 1 & 1\\
    M3 & 1 & 1\\
    M4 & 1 & 1\\
    M5 & 0 & 0\\
    M6 & 0 & 1\\
    M7 & 0 & 1\\
    M8 & 1 & 0\\
\noalign{\smallskip}\hline
  \end{tabular}
}   
\subfloat{
 \begin{tabular}{ @{} rcl@{} }
&&\\
&&\\
&&\\
TPR&=&0.8\\
FPR&=& 0.66\\
PPV&=& 0.66 \\
NPV&=& 0.5 \\
OA&=& 0.625 \\
FN/FP&=& 0.5\\
 \end{tabular} 
  }
\end{table}

\textcolor{black}{
Treatment equality can be suitable to use in case the cost (or benefit) of a FP is a fixed ratio (or reciprocal) of the cost (or benefit) of a FN. For example, one can think of a loan granting scenario where the cost of a FP (misclassifying a non-defaulter) is exactly a fraction (e.g. 1/3) of the cost of a FN (misclassifying a defaulter).}

\textbf{Total fairness}~\citep{berk2018fairness} is another notion which holds when all aforementioned fairness notions are satisfied simultaneously, that is, statistical parity, equalized odds, conditional use accuracy equality (hence, overall accuracy equality), and treatment equality. Total fairness is a very strong notion which is very difficult to hold in practice. Table~\ref{tab:tf} shows a scenario where total fairness holds. More generally, total fairness is satisfied in the very uncommon situation where the proportions of TPs, TNs, FPs, and FNs are the same in all groups.    

\begin{table}[!h]
	\centering
	\caption{A job hiring scenario satisfying total fairness. \medskip}
\label{tab:tf} 
\subfloat{
 \begin{tabular}{ @{} rcl@{} }
&&\\
&&\\
&&\\
TPR&=&0.5\\
FPR&=& 0.66\\
PPV&=& 0.33 \\
NPV&=& 0.5 \\
OA&=& 0.4 \\
FN/FP&=& 0.5\\
  \end{tabular} 
  }
  \captionsetup[subfloat]{position=top,labelformat=empty}
\subfloat[Group 1 (Female)]{
 \begin{tabular}{ @{} ccc@{} }
\hline\noalign{\smallskip}
Gender & $\; Y \; $    & $\; \hat{Y} \; $ \\
    \noalign{\smallskip}\hline\noalign{\smallskip}
    F1 & 1 & 1\\
    F2 & 0 & 0\\
    F3 & 0 & 1\\
    F4 & 0 & 1\\
    F5 & 1 & 0\\
     &  & \\
     &  & \\
     &  & \\
     &  & \\
     &  & \\
   \noalign{\smallskip}\hline
  \end{tabular}  
  }
 \qquad 
 \subfloat[Group 2 (Male)]{
   \begin{tabular}{ @{} ccc@{} }
\hline\noalign{\smallskip}
Gender & $\; Y \; $ & $\; \hat{Y} \; $  \\
\noalign{\smallskip}\hline\noalign{\smallskip}
    M1 & 1 & 1\\
    M2 & 1 & 1\\
    M3 & 0 & 0\\
    M4 & 0 & 0\\
    M5 & 0 & 1\\
    M6 & 0 & 1\\
    M7 & 0 & 1\\
    M8 & 0 & 1\\
    M9 & 1 & 0\\
    M10 & 1 & 0\\
\noalign{\smallskip}\hline
  \end{tabular}
}   
\subfloat{
 \begin{tabular}{ @{} rcl@{} }
&&\\
&&\\
&&\\
TPR&=&0.5\\
FPR&=& 0.66\\
PPV&=& 0.33 \\
NPV&=& 0.5 \\
OA&=& 0.4 \\
FN/FP&=& 0.5\\
 \end{tabular} 
  }
\end{table}

\textcolor{black}{
Total fairness can be considered in scenarios where any deviation in misclassification or acceptance rates between sub-populations is very costly\footnote{The cost can be financial, ethical, reputation, etc.}.
}

\subsection{Balance}
 \label{sec:balance}

 The predicted outcome ($\hat{Y}$) is typically derived from a score ($S$) which is returned by the ML algorithm. All aforementioned fairness notions do not use the score to assess fairness. Typically, the score value is normalized to be in the interval $[0,1]$ which makes it possible to interpret the score as the probability to predict the sample as positive. \textbf{Balance for positive class}~\citep{kleinberg_et_al:LIPIcs:2017:8156} focuses on the applicants who constitute positive instances and is satisfied if the average score $S$ received by those applicants is the same for both groups. 
 In other words, a violation of this balance means that applicants belonging to the positive class in one group might receive steadily lower predicted score than applicants belonging to the positive class in the other group:
\begin{equation}
\label{eq:balPosclass}
E[S \mid Y =1,A = 0)] = E[S \mid Y =1,A = 1] \end{equation}

Table~\ref{tab:bal} shows a job hiring scenario where the average score for female candidates that should be hired ($Y=1$) is $7.1$ while it is $4.7$ for male candidates. The scenario is not balanced for positive class. Note that, despite the significant difference between these two average values, for a score threshold value of $5$, the scenario of Table~\ref{tab:bal} satisfies both statistical parity (Eq.~\ref{eq:sp}) and equal opportunity (Eq.~\ref{eq:eqOpp}).

\begin{table}[!h]
	\centering
	\caption{A job hiring scenario satisfying statistical parity and equal opportunity (for a score threshold value of $5$) but neither balance for positive class nor balance for negative class. \medskip}
\label{tab:bal} 
\subfloat[Group 1 (Female)]{
 \begin{tabular}{ @{} ccc@{} }
\hline\noalign{\smallskip}
    Gender & $Y$    & $S$ \\
    \noalign{\smallskip}\hline\noalign{\smallskip}
    F1 & 1 & 9\\
    F2 & 1 & 8\\
    F3 & 0 & 8\\
    F4 & 1 & 4.5\\
    F5 & 0 & 4.5\\
    F6 & 0 & 3.5\\
   \noalign{\smallskip}\hline
  \end{tabular}  
  }
 \qquad 
 \subfloat[Group 2 (Male)]{
   \begin{tabular}{ @{} ccc@{} }
\hline\noalign{\smallskip}
    Gender & $Y$ & $S$  \\
\noalign{\smallskip}\hline\noalign{\smallskip}
    M1 & 1 & 6.2\\
    M2 & 1 & 6\\
    M3 & 0 & 5.5\\
    M4 & 0 & 1\\
    M5 & 1 & 2\\
    M6 & 0 & 2\\
\noalign{\smallskip}\hline
  \end{tabular}
}   
\end{table}

\textbf{Balance of negative class}~\citep{kleinberg_et_al:LIPIcs:2017:8156} is an analogous fairness notion where the focus is on the negative class:
\begin{equation}
\label{eq:balNegclass}
E[S \mid Y =0,A = 0] = E[S \mid Y =0,A = 1]
\end{equation}

The scenario in Table~\ref{tab:bal} is not balanced for the negative class either since the average scores for the negative class ($Y=0$) for the female and male groups are $5.3$ and $2.8$, respectively. 

Both variants of balance can be required simultaneously (Eq.~\ref{eq:balPosclass} and~\ref{eq:balNegclass}) which leads to a stronger notion of balance. Since no previous work reported such fairness notion, for completeness, we define it and call it \textbf{overall balance}.

\begin{definition}
Overall balance is satisfied iff:
\begin{equation}
\label{eq:ovBal}
E[S \mid Y =y,A = 0] = E[S \mid Y =y,A = 1] \quad \forall y \in \{0,1\}
\end{equation}
\end{definition}

Balance fairness notions are relevant in the criminal risk assessment scenario because a divergence in the score values of individuals from different races may indicate a difference in the type of crime that can be committed (high risk score typically means a serious crime). \textcolor{black}{Balance fairness notions are also suitable in the teacher firing scenario since any discrepancy between the average evaluation scores of fired teachers in different groups is a clear indicator of bias. On the other hand, balance fairness notions can be misleading in presence of clusters of samples sharing very similar attribute values and having score values in the vicinity of the positive/negative outcome threshold. In such case, the average score of the positive/negative class can change significantly due to a slight increase/decrease of the threshold value.}

\subsection{Calibration}
 \label{sec:calib}

 Calibration \citep{chouldechova2017fair} (a.k.a. test-fairness \citep{chouldechova2017fair}, matching conditional frequencies \citep{hardt2016equality}) relies on the score variable as follows. To satisfy calibration, for each predicted probability score $S=s$, 
individuals in all groups should have the same probability to actually belong to the
positive class: 
\begin{equation}
\label{eq:calib}
P(Y =1 \mid S =s,A = 0) = P(Y =1 \mid S =s,A = 1) \quad \forall s \in [0,1]
\end{equation}

Eq.~\ref{eq:calib} is very unlikely to be satisfied in practice as the probability of two individuals having exactly the same real number score is very small. Moreover, technically, the probability that $S$ exactly equal to $s$ is typically $0$. Therefore, in practice, the space of score values $[0,1]$ is binned into intervals called bins such that any two values falling in the same bin are considered equal~\citep{kleinberg_et_al:LIPIcs:2017:8156, verma2018fairness,kleinberg2018human}.

In our job hiring example, this implies that for any score value $s\in[0,1]$, the probability of truly being hired should be the same for both male and female applicants.

\begin{table}[!h]
	\centering
	\caption{A job hiring scenario satisfying predictive parity (for any threshold smaller than $0.7$ or larger than $0.8$) but not calibration. \medskip}
\label{tab:cal} 
\subfloat[Group 1 (Female)]{
 \begin{tabular}{ @{} ccc@{} }
\hline\noalign{\smallskip}
    Gender & $Y$    & $S$ \\
    \noalign{\smallskip}\hline\noalign{\smallskip}
    F1 & 1 & 0.85\\
    F2 & 1 & 0.8\\
    F3 & 0 & 0.8\\
    F4 & 1 & 0.7\\
    F5 & 0 & 0.7\\
    F6 & 0 & 0.4\\
    F7 & 1 & 0.4\\
    F8 & 0 & 0.4\\
   \noalign{\smallskip}\hline
  \end{tabular}  
  }
 \qquad 
 \subfloat[Group 2 (Male)]{
   \begin{tabular}{ @{} ccc@{} }
\hline\noalign{\smallskip}
    Gender & $Y$ & $S$  \\
\noalign{\smallskip}\hline\noalign{\smallskip}
    M1 & 1 & 0.85\\
    M2 & 1 & 0.8\\
    M3 & 1 & 0.8\\
    M4 & 0 & 0.7\\
    M5 & 0 & 0.7\\
    M6 & 1 & 0.4\\
    M7 & 0 & 0.4\\
    M8 & 0 & 0.4\\
\noalign{\smallskip}\hline
  \end{tabular}
}   
\end{table}
\normalsize

Eq.~\ref{eq:calib} is very similar to Eq.~\ref{eq:predPar} corresponding to predictive parity. Table~\ref{tab:cal} illustrates a job hiring scenario that may or may not satisfy predictive parity depending on the score threshold to hire a candidate; for a threshold value of $0.6$, PPV rate for both male and female groups is the same, $0.6$, while for a threshold value of $0.75$, PPV for female group is $0.66$ but for male it is $1.0$. However, the calibration score ($P(Y =1 \mid S =s,A = a)\;\; a\in\{0,1\}$\textcolor{black}{,\; $s\in[0,1]$})  for every value of $s$ is as follows:
\begin{center}
\begin{tabular}{lllll}
\hline\noalign{\smallskip}
s & 0.4 & 0.7 & 0.8 & 0.85\\
\noalign{\smallskip}\hline\noalign{\smallskip}
Female        & 0.33 & 0.5 & 0.5 & 1.0\\ 
Male          & 0.33 & 0   & 1.0 & 1.0 \\ 
 \noalign{\smallskip}\hline
\end{tabular}
\end{center}
Calibration is satisfied for score values $0.4$ and $0.85$, but not satisfied for score values $0.7$ and $0.8$. Overall, the scenario of Table~\ref{tab:cal} does not satisfy calibration.

\begin{table}[!h] \color{black}
	\centering
	\caption{\color{black} A job hiring scenario satisfying calibration but not predictive parity (for any threshold). \medskip}
\label{tab:cal2} 
\subfloat[Group 1 (Female)]{
 \begin{tabular}{ @{} ccc@{} }
\hline\noalign{\smallskip}
    Gender & $Y$    & $S$ \\
    \noalign{\smallskip}\hline\noalign{\smallskip}
    F1 & 1 & 0.8\\
    F2 & 1 & 0.8\\
    F3 & 1 & 0.7\\
    F4 & 1 & 0.7\\
    F5 & 0 & 0.7\\
    F6 & 0 & 0.7\\
    F7 & 0 & 0.3\\
    F8 & 0 & 0.3\\
   \noalign{\smallskip}\hline
  \end{tabular}  
  }
 \qquad 
 \subfloat[Group 2 (Male)]{
   \begin{tabular}{ @{} ccc@{} }
\hline\noalign{\smallskip}
    Gender & $Y$ & $S$  \\
\noalign{\smallskip}\hline\noalign{\smallskip}
     &  & \\
    M1 & 1 & 0.8\\
    M2 & 1 & 0.8\\
    M3 & 1 & 0.7\\
    M4 & 0 & 0.7\\
    M5 & 0 & 0.3\\
    M6 & 0 & 0.3\\
    &  & \\
\noalign{\smallskip}\hline
  \end{tabular}
}   
\end{table}

\textcolor{black}{Interestingly, calibration is not always stronger than predictive parity~\cite{garg2020fairness}. Table~\ref{tab:cal2} shows a job hiring scenario satisfying calibration, but not predictive parity.}
Calibration is suitable to use in scenarios where the threshold is not fixed and is very likely to be tuned to accommodate a changing context. A first example is the acceptance score in loan granting applications which may change abruptly due to economic instability. A second example is the child maltreatment risk assessment where the threshold for intervention (withdrawing a child from his family) depends on the available seats in foster houses. 

\textbf{Well-calibration}~\citep{kleinberg_et_al:LIPIcs:2017:8156} is a stronger variant of calibration. It requires that (1) calibration is satisfied, (2) the score is interpreted as the probability to truly belong to the positive class, and (3) for each score $S=s$, the probability to truly belong to the positive class is equal to that particular score: 

\begin{equation}
\label{eq:wellCalib}
P(Y =1 \mid S =s,A = 0) = P(Y =1 \mid S =s,A = 1) = s  \quad  \forall \; {s \in [0,1]}
\end{equation}
Intuitively, for a set of applicants who have a certain probability $s$ of being hired, approximately $s$ percent of these applicants should truly be hired. Table~\ref{tab:welcal} (a) is a job hiring scenario which is calibrated (the proportion of applicants which should be hired for every score value is the same for male and female groups) but not well-calibrated (the score value does not coincide with the proportion of applicants that should be hired). Table~\ref{tab:welcal} (b) is both calibrated and well-calibrated. Garg et al.~\citep{garg2020fairness} show that the difference between calibration and well-calibration is a simple difference in mapping. That is, ``the scores of a calibrated predictor can, using a suitable transformation, be converted to scores satisfying well-calibration''.

\begin{table}[!h]
	\centering
	\caption{Calibration vs well-calibration. \medskip}
\label{tab:welcal} 
\subfloat[Calibrated but not well-calibrated]{
\begin{tabular}{lllll}
\hline\noalign{\smallskip}
s & 0.4 & 0.7 & 0.8 & 0.85\\
\noalign{\smallskip}\hline\noalign{\smallskip}
Female        & 0.33 & 0.5 & 0.6 & 0.6\\ 
Male          & 0.33 & 0.5   & 0.6 & 0.6 \\ 
 \noalign{\smallskip}\hline
\end{tabular}
\ }
 \qquad 
 \subfloat[Calibrated and well-calibrated]{
\begin{tabular}{lllll}
\hline\noalign{\smallskip}
s & 0.4 & 0.7 & 0.8 & 0.85\\
\noalign{\smallskip}\hline\noalign{\smallskip}
Female        & 0.4 & 0.7 & 0.8 & 0.85\\ 
Male          & 0.4 & 0.7   & 0.8 & 0.85 \\ 
 \noalign{\smallskip}\hline
\end{tabular}
}   
\end{table}

\subsection{Group vs individual fairness notions}

All the fairness notions discussed above are considered as group fairness where their common objective is to ensure that groups who differ by their sensitive attributes are treated equally. These notions, mainly based on statistical measures, generally ignore all attributes of the individuals except the sensitive attribute $A$. Such treatment might hide unfairness. Dwork et al.~\citep{dwork2012fairness} stated that group fairness, despite its suitability for policies among demographic sub-populations, does not guarantee that individuals are treated fairly. This is illustrated in the simple example in Table~\ref{tab:IndvsGr}. The example satisfies most of group fairness notions, including total fairness (Section~\ref{sec:treatEq}). However, based on the applicants profiles, it is clear that the predictor is unfair towards applicant Female 4. The fairness notions which follow  attempt to address such issues by not marginalizing over non-sensitive attributes $X$ of an individual, therefore they are called individual fairness notions~\footnote{The term individual fairness is used in some papers to refer to fairness through awareness (Section~\ref{sec:FTA}). In this paper, the term individual fairness refers to fairness notions which cannot be considered as group fairness notions.}. 

\begin{table}[!h]
	\centering
	\caption{A simple job hiring example satisfying most of group fairness notions, but unfair towards Female 4 applicant. \medskip}
\label{tab:IndvsGr} 
\subfloat{
 \begin{tabular}{ @{} cLMcLcc@{} }
\hline\noalign{\smallskip}
Gender    & Education Level & Job Experience &  Age  & Marital Status & Y & $\hat{Y}$\\
    \noalign{\smallskip}\hline\noalign{\smallskip}
    Female 1           & 8    & 2 & 39            & single    & 0 & 1 \\
   Female 2            & 8     & 2   & 26             & married & 0 & 1 \\
    Female 3            & 6     & 1   & 32             & married & 0  & 0\\
    Female 4     & 12      &  8   & 35             & single  & 1   & 0  \\
    Female 5     & 9      &  10   & 29             & married  & 1   & 1 \\
\hline 
    Male 1          & 7    & 3 & 34            & single    & 0 & 1  \\
    Male 2          & 8     & 0   & 28             & married &  1  & 0 \\
    Male 3        & 11     & 8   & 43             & single     & 1 & 1   \\
    Male 4      & 7      &  1   & 26             & married     & 0 & 0 \\
    Male 5          &8     & 2   & 41             & single   & 0 & 1 \\
\noalign{\smallskip}\hline
  \end{tabular}  
  }
 \qquad 
\subfloat{
 \begin{tabular}{ @{} rcl@{} }
&&\\
&&\\
&&\\
TPR&=&0.5\\
FPR&=&0.66\\
PPV&=&0.33\\
OA&=&0.4\\
&&\\
TPR&=& 0.5 \\
FPR&=& 0.66 \\
PPV&=&0.33\\
OA&=& 0.4 \\
&&\\
 \end{tabular} 
  }
\end{table}

\subsection{Causal discrimination}
\label{sec:causDiscrim}
Causal Discrimination~\citep{galhotra2017fairness} implies that a classifier should produce exactly the same prediction for individuals who differ only from gender  while possessing identical attributes X. In our hiring example, this is to say that male and female applicants with the same attributes X should have the same predictions:

\begin{equation}
\label{eq:cd}
X\textsubscript{\textcolor{black}{(A=0)}} = X \textsubscript{\textcolor{black}{(A=1)}}  \; \land \; A\textsubscript{\textcolor{black}{(A=0)}} \; \neq  A\textsubscript{\textcolor{black}{(A=1)}} \; \Rightarrow \hat{y}\textsubscript{\textcolor{black}{(A=0)}} =  \hat{y}\textsubscript{\textcolor{black}{(A=1)}}
\end{equation}
In our example, this implies that male and female applicants who otherwise have the same attributes X will either both be assigned a positive prediction or both assigned a negative prediction. Considering the example of Table~\ref{tab:example1}, two applicants of different genders (Female 2 and Male 4) have identical values of X yet, getting different predictions (negative for female applicant while positive for male applicant). The predictor is then unfair towards Female 2 applicant. 

At a first glance, causal discrimination can be seen as an extreme case of conditional statistical parity (Section~\ref{sec:notioncsp}) when conditioning on all non-sensitive attributes ($E = X$). However, conditional statistical parity is a group fairness notion which is satisfied if the proportion of individuals having the same non-sensitive attribute values and predicted accepted in both groups (e.g. male and female) is the same. This is why Eq.~\ref{eq:csp} is expressed in terms of conditional probabilities. Causal discrimination, however, consider every individual separately regardless of its contribution to sub-population proportions. To illustrate this subtlety, consider the following scenario:

\begin{center}
\begin{tabular}{ @{} cLMcLcc@{} }
    \noalign{\smallskip}\hline\noalign{\smallskip}
    Female 1           & 8    & 2 & 26            & single    & $\hat{Y}=0$ &  \\
   Female 2            & 8     & 2   & 26             & single & $\hat{Y}=1$ &  \\
\hline 
    Male 1          & 8    & 2 & 26            & single    & $\hat{Y}=1$ &   \\
    Male 2          & 8     & 2   & 26             & single &  $\hat{Y}=0$ &  \\
\noalign{\smallskip}\hline
  \end{tabular} 
\end{center}

Conditional statistical parity with $E=X$ (conditioning on all non-sensitive attributes) is satisfied as the proportion of males and females having the exact same attribute values and predicted accepted is the same ($0.5$). However, at the individual level, causal discrimination is not satisfied as there are two violations: Female 1 vs Male 1 and Female 2 vs Male 2. The two violations compensated each others and as a result conditional statistical parity is satisfied. 

Causal discrimination is suitable to use in decision making scenarios where it is very common to find individuals sharing exactly the same attribute values. For example, admission decision making based mainly on test scores and categorical attributes. To apply this fairness notion on a loan granting scenario where there are only few individuals with exactly the same attribute values, Verma and Rubin~\citep{verma2018fairness} generated, for every applicant in the dataset, an identical individual of the opposite gender. The result of applying causal discrimination is the percentage of violations in the entire population (i.e. how many individuals are unfairly treated?).

\subsection{Fairness through awareness}
\label{sec:FTA}
Fairness through awareness~\citep{dwork2012fairness} (a.k.a individual fairness~\citep{gajane2017survey,kusner2017counterfactual}) is a generalization of causal discrimination which implies that similar individuals should have similar predictions. Let $i$ and $j$ be two individuals represented by their attributes values vectors $v_i$ and $v_j$. Let $d(v_i,v_j)$ represent the similarity distance between individuals $i$ and $j$. Let $M(v_i)$ represent the probability distribution over the outcomes of the prediction. For example, if the outcome is binary ($0$ or $1$), $M(v_i)$ might be $[0.2,0.8]$ which means that for individual $i$, $P(\hat{Y}=0) = 0.2$ and $P(\hat{Y}=1) = 0.8$. Let $D$ be a distance metric between probability distributions.  
Fairness through awareness is achieved iff, for any pair of individuals $i$ and $j$:
\begin{equation}
\label{eq_FTA}
D(M(v_i), M(v_j))  \leq d(v_i, v_j)
\end{equation} 

For our hiring example, this implies that the distance between the distribution of outcomes of two applicants should be at most the distance between those applicants\footnote{Reducing all difference between two applicants/instances to a single distance value is often not easy to do in practice.}. A possible relevant features to use for measuring the similarity between two applicants might be the education level and the job experience. The distance metric $d$ between two applicants could be defined as the average of the normalized difference (the difference divided by the maximum difference in a dataset) of their education level and their job experience. More formally, let $E_{v_i}$ and $E_{v_j}$  be the education levels of individuals $i$ and $j$, respectively, and let $N_E$ be the normalized difference between education levels, that is,  $N_E= \frac{\mid E_{v_i} - E_{v_j} \mid}{m_E}$ where $m_E$ is the maximum difference in education level in the dataset. Similarly, let $J_{v_i}$ and $J_{v_j}$  be the job experience of individuals $i$ and $j$, while $N_J$ is the normalized difference of the job experience, that is, $N_J = \frac{\mid J_{v_i} - J_{v_j} \mid}{m_J} $ where $m_J$ is the maximum difference in job experience in the dataset. The distance metric is defined as:
\[
  d(v_i, v_j) = \frac{N_E + N_J}{2},
\]

The distance between the probability distributions over the outcomes could be the \textit{Hellinger distance}~\citep{nikulin2001hellinger}. Let $\{y_1,y_2,\ldots,y_K\}$ be the set of possible outcomes and let $P$ and $Q$ two (discrete) probability distributions. The Hellinger distance between $P$ and $Q$ is defined as:
\[ H(P,Q) = \frac{1}{\sqrt{2}} \sqrt{\sum_{k=1}^{K} \left(\sqrt{P(y_k)} - \sqrt{Q(y_k)} \right)^2} \]
Table~\ref{tab:FTA} shows a sample from the job hiring dataset on which fairness through awareness is applied. The result of applying fairness through awareness is shown in Table~\ref{tab:FTAapplied}. Each cell at the left of the shaded diagonal represents a distance between two individuals and each cell at the right of the shaded diagonal represents the distance between probability outcomes of two individuals. \newline
For instance: \[d(F1, F2) = 0.25\]
While:
\begin{align}
     D(M(F1), M(F2))  =  \;\;&  \frac{1}{\sqrt{2}} \sqrt{\left(\sqrt{0.4} - \sqrt{0.3} \right)^2 + \left(\sqrt{0.6} - \sqrt{0.7} \right)^2} \nonumber\\
    = \;\; & \frac{1}{\sqrt{2}} \sqrt{ 0.0081 + 0.0036} \nonumber\\
    = \;\; & 0.07 \nonumber
\end{align}


The cell values in bold represent the cases where fairness through awareness is not satisfied: $D  \nleq d$. For example, $\textbf{0.07}$ ($< 0.0$) implies that $F_1$ is discriminated compared to  $M_3$. Similarly, $M_2$ is discriminated compared to $F_3$, $F_2$, and $M_3$.

\begin{table}[!h]
\centering
\caption{Job hiring sample used to apply fairness through awareness. \medskip}
\label{tab:FTA} 
\subfloat[Dataset]{
 \begin{tabular}{ @{} cLMcLc@{} }
\hline\noalign{\smallskip}
    Gender    & Education Level & Job Experience &  Age  & Marital Status & Label \\
    \noalign{\smallskip}\hline\noalign{\smallskip}
    Female 1           & 12    & 2 &39            & single    & 1 \\
   Female 2            & 12     & 1   & 26      & married     & 0\\
   Female 3            & 13     & 1   & 32      & married     & 1\\
   Male 1      & 13      &  1   & 26             & married     & 1 \\
    Male 2          &12     & 1   & 41             & single     & 0  \\
    Male 3          &12     & 2   & 30             & single     & 1  \\
   \noalign{\smallskip}\hline
  \end{tabular}  
  }
 \qquad 
  \subfloat[Prediction]{
   \begin{tabular}{ @{} NN@{} }
\hline\noalign{\smallskip}
    \^{Y} & S\  \\
    \\
\noalign{\smallskip}\hline\noalign{\smallskip}
    0 & 0.4\\
    0 & 0.3\\
    1 & 0.9\\
    0 & 0.2\\
    0 & 0.2\\
    1 & 0.7\\
\noalign{\smallskip}\hline
  \end{tabular}
}   
\end{table}

\begin{table}[!h]
\caption{Application of fairness through awareness. Each cell at the left of the shaded table's diagonal represents a distance between a pair of applicants. Those at the right represent the distance between probability distributions. Values in bold imply cases where $D$ > $d$, meaning fairness through awareness is not satisfied.
 \medskip}
\centering
\label{tab:FTAapplied}    
  \begin{tabular}{|l|W|W|c|W|W|W|l}
\cline{1-7}
 & F1 & F2 & F3 & M1 & M2 & M3 &{\small{\multirow{7}{*}{\rotatebox[origin=c]{90}{$D(M(v_i),M(v_j))$}}}} \\
\cline{1-7}
   F1            & {\cellcolor{gray}}& 0.07&0.26 & 0.16& 0.16 & \textbf{0.07} \\
\cline{1-7}
    F2            & 0.25 & {\cellcolor{gray}}& 0.18& 0.08 & \textbf{0.08} & \textbf{0.29}\\
\cline{1-7}
    F3             & 0.75  & 0.5&{\cellcolor{gray}} &0.1 & \textbf{0.54}& 0.18\\
\cline{1-7}
    M1             & 0.75 & 0.5&0.0 & {\cellcolor{gray}}&0.0 & 0.08\\
\cline{1-7}
    M2        &  0.25& 0.0&0.5 &0.5 &{\cellcolor{gray}} & \textbf{0.37}\\
\cline{1-7}
   M3            &0.0  & 0.25&0.75 &0.75 & 0.25& {\cellcolor{gray}}\\
\cline{1-7}
\multicolumn{1}{l}{}&\multicolumn{1}{l}{} & \multicolumn{1}{l}{}& \multicolumn{1}{l}{$d(v_i,v_j)$}&\multicolumn{1}{l}{} & \multicolumn{1}{l}{}&\multicolumn{1}{l}{} \\
\end{tabular}
\end{table}

Fairness through awareness is more fine-grained than any group fairness notion presented earlier in Sections~\ref{sec:notion_sp}–~\ref{sec:calib}. For instance, in the example of Table~\ref{tab:FTA}, statistical parity is satisfied: $0.33$ for both men and women. Likewise, equalized odds (~\ref{sec:eqOdds}) is satisfied as the TPR and the FPR are equal for male and female applicants ($0.5$ and $0$, respectively). Nevertheless, Table~\ref{tab:FTAapplied} shows that when comparing each pair of individuals (regardless of their gender) cases of discrimination have been discovered.

\begin{figure}[!h]
\centering
\includegraphics[scale=0.27]{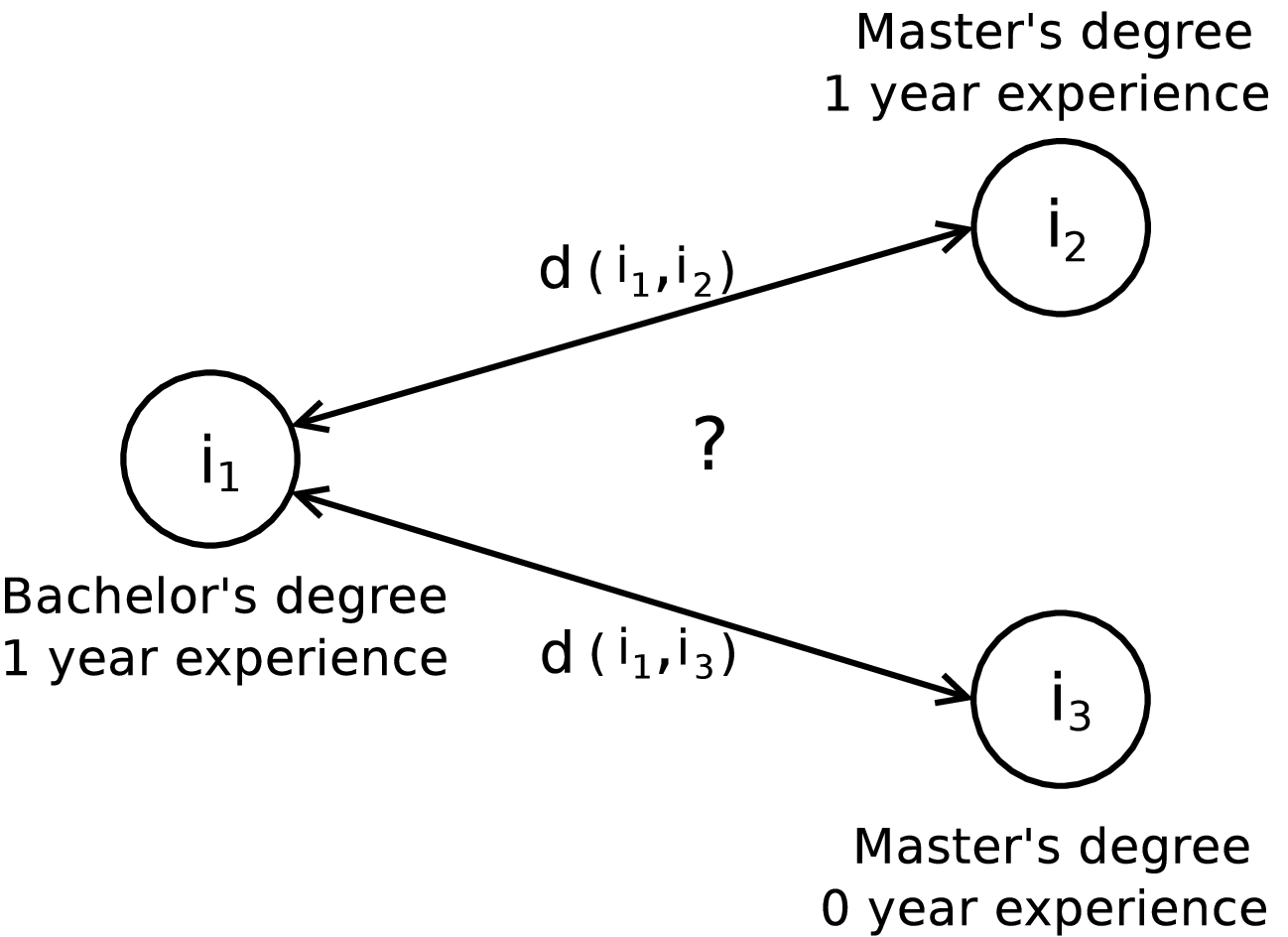}
\caption\small{{An example showing the difficulty of selecting a distance metric in fairness through awareness}}
\label{fig:fta}     
\end{figure}

It is important to mention that, in practice, fairness through awareness introduces some challenges. For instance, it assumes that the similarity metric is known for each pair of individuals~\citep{kim2018fairness}. That is, a challenging aspect of this approach is the difficulty to determine what is an appropriate metric function to measure the similarity between two individuals. Typically, this requires careful human intervention from professionals with domain expertise~\cite {kusner2017counterfactual}. For instance, suppose a company is intending to hire only two employees while three applicants $i_1$, $i_2$ and $i_3$ are eligible for the offered job. Assume $i_1$ has a bachelor’s degree and $1$ year related work experience, $i_2$ has a master’s degree and $1$ year related work experience and $i_3$ has a master’s degree but no related work experience (Figure~\ref{fig:fta}). Is $i_1$ closer to $i_2$ than $i_3?$ If so, by how much? This is difficult to answer, especially if the company overlooked such specific cases and did not carefully define and set a suitable and fair similarity metric in order to rank applicants for job selection. Thus, fairness through awareness can not be considered suitable for domains where trustworthy and fair distance metric is not available.

{\color{black}
\subsection{Causality-based fairness notions}
\label{sec:causalNotions}

Causality-based fairness notions differ from all aforementioned statistical fairness approaches in that they are not totally based on data but consider additional knowledge about the structure of the world, in the form of a causal model. This additional knowledge helps us understand how data is generated in the first place and how changes in variables propagate in a system. Most of these fairness notions are defined in terms of non-observable quantities such as interventions (to simulate random experiments) and counterfactuals (which consider other hypothetical worlds, in addition to the actual world).

A variable $X$ is a cause of a variable $Y$ if $Y$ in any way relies on $X$ for its value~\citep{pearl2016book}. Causal relationships are expressed using structural equations~\citep{bollen1989structural} and represented by causal graphs where nodes represent variables (attributes) and edges represent causal relationships between variables. Figure~\ref{fig:CDCounterfactual} shows a possible causal graph for our hiring example where directed edges indicate causal relationships.  

\begin{figure}[!h]
\centering
\includegraphics [scale=0.27] {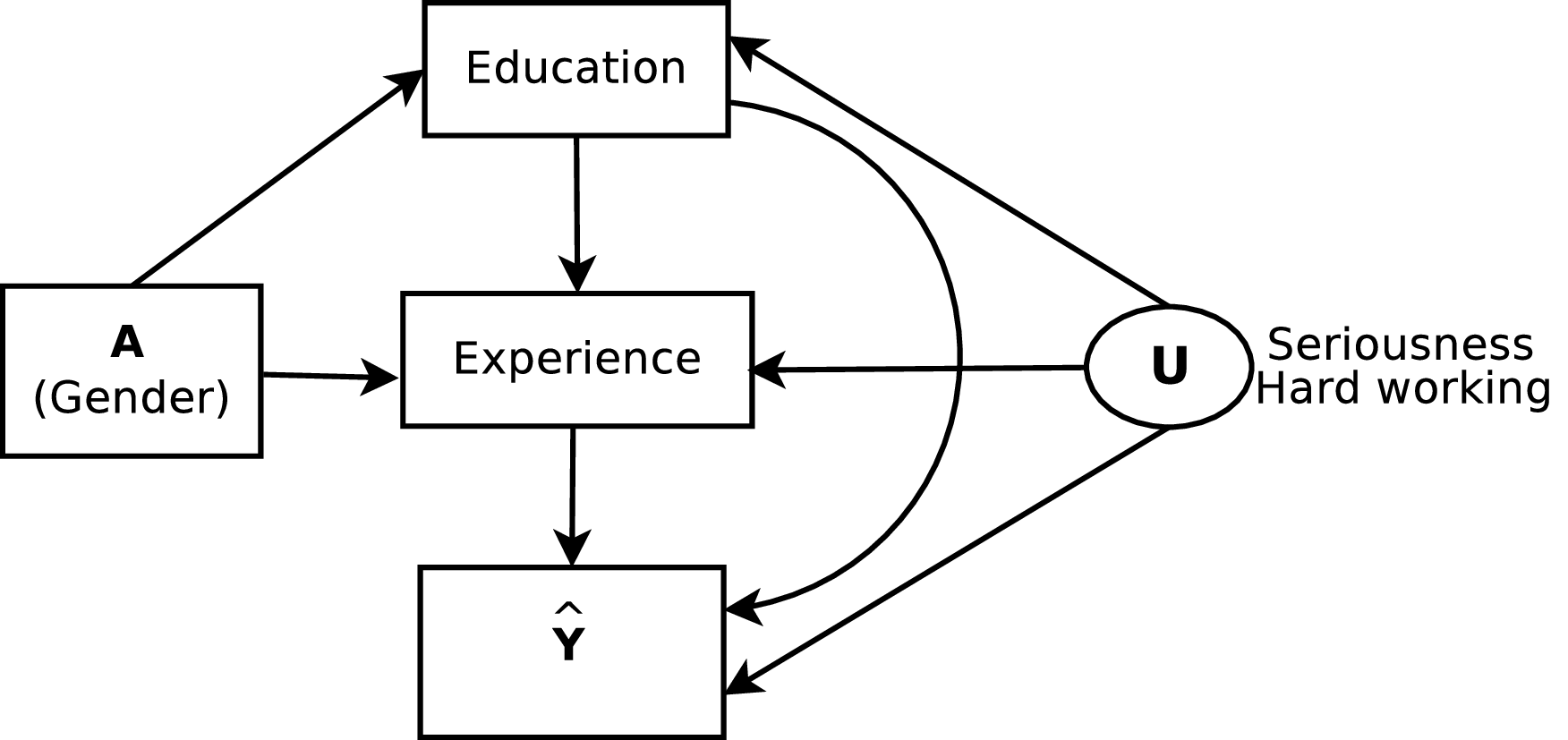}
\caption{A possible causal graph for the hiring example.}
\label{fig:CDCounterfactual}     
\end{figure}

Statistical parity (Section~\ref{sec:notion_sp}) is known also as total variation (TV) as it can be expressed by subtracting the two terms in Eq.~\ref{eq:sp} as follows:
\begin{equation}
\label{eq:TV}
TV_{a_1,a_0} (\hat{y}) = P(\hat{Y} = \hat{y} \mid A=a_1) - P(\hat{Y} = \hat{y} \mid A=a_0)
\end{equation}
A $TV$ equal zero indicates fairness according to statistical parity.
As $TV$ is purely a statistical notion, it is unable to reflect the causal relationship between $A$ and $Y$, that is, it is insensitive to the mechanism by which data is generated.  

Total effect ($TE$)~\citep{pearl2009causality} is the causal version of $TV$ and is defined in terms of experimental probabilities as follows : 
\begin{equation}
\label{eq:TE}
TE_{a_1,a_0} (\hat{y}) = P(\hat{y}_{A\leftarrow a_1}) - P(\hat{y}_{A\leftarrow a_0})
\end{equation}

$P(\hat{y}_{A\leftarrow a}) = P(\hat{Y}=\hat{y} \mid do(A=a))$ is called an experimental probability and is expressed using intervention. An intervention, noted $do(V=v)$, is a manipulation of the model that consists in fixing the value of a variable (or a set of variables) to a specific value. Graphically, it consists in discarding all edges incident to the vertex corresponding to variable $V$.  Intuitively, using the job hiring example, while $P(\hat{Y}=1\mid A=0)$ reflects the probability of hiring among female applicants, $P(\hat{Y}_{A\leftarrow 0}=1 = P(\hat{Y}=1) \mid do(A=0))$ reflects the probability of hiring if \textit{all the candidates in the population} had been female. The obtained distribution $P(\hat{Y}_{A\leftarrow a})$ can be considered as a \textit{counterfactual} distribution since the intervention forces $A$ to take a value different from the one it would take in the actual world. Such counterfactual variable is noted also $\hat{Y}_{A=a}$ or $\hat{Y}_a$ for short.

$TE$ measures the effect of the change of $A$ from $a_1$ to $a_0$ on $\hat{Y}=\hat{y}$ along all the causal paths from $A$ to $\hat{Y}$. 
Intuitively, while $TV$ reflects the difference in proportions of $\hat{Y}=\hat{y}$ in the current cohort, $TE$ reflects the difference in proportions of $\hat{Y}=\hat{y}$ in the entire population.
A more involved causal-based fairness notion considers the effect of a change in the sensitive attribute value (e.g. gender) on the outcome (e.g. probability of hiring) given that we already observed the outcome for that individual. This typically involves an impossible situation which requires to go back in the past and change the sensitive attribute value. Mathematically, this can be formalized using counterfactual quantities. The simplest fairness notion using counterfactuals is the effect of treatment on the treated (ETT)~\citep{pearl2009causality}.


The effect of treatment on the treated (ETT) is defined as:
\begin{equation}
\label{eq:ETT}
ETT_{a_1,a_0} (\hat{y}) = P (\hat{y}_{A\leftarrow a_1} \mid a_0) - P (\hat{y} \mid a_0)
\end{equation}
$P(\hat{y}_{A\leftarrow a_1}\mid a_0)$ reads the probability of $\hat{Y}=\hat{y}$ had $A$ been $a_1$, given $A$ had been observed to be $a_0$. For instance, in the job hiring example, $P(\hat{Y}_{A\leftarrow 1} \mid A=0)$ reads the probability of hiring an applicant had she been a male, given that the candidate is observed to be female. Such probability involves two worlds: an actual world where $A=a_0$ (the candidate is female) and a counterfactual world where for the same individual $A=a_1$ (the same candidate is male). Notice that $P(\hat{y}_{A\leftarrow a_0}\mid a_0) = P(\hat{y} \mid a_0)$, a property called consistency~\citep{pearl2009causality}.

Counterfactual fairness~\citep{kusner2017counterfactual} is a fine-grained variant of ETT conditioned on all attributes. That is, a prediction $\hat{Y}$ is counterfactually fair if under any assignment of values $X=x$,
\begin{equation}
\label{eq:counterfactual}
P(\hat{Y}_{A\leftarrow a_1}=\hat{y} \mid X=x, A = a_0) = P(\hat{Y}_{A\leftarrow a_0}=\hat{y} \mid X=x, A = a_0) 
\end{equation}
where $X$ is the set of all attributes excluding $A$. 
Since conditioning is done on all remaining variables $X$, counterfactual fairness is an individual notion. According to Eq.~\ref{eq:counterfactual}, counterfactual fairness is satisfied if the probability distribution of the outcome $\hat{Y}$ is the same in the actual and counterfactual worlds, for every possible individual. In the job hiring example, an MLDM is counterfactually fair if:
\begin{equation}
\label{eq:counterfactualJob}
P(\hat{Y}_{A\leftarrow 1} \mid X=x, A = 0) = P(\hat{Y}_{A \leftarrow 0} \mid X=x, A = 0) 
\end{equation}

The main problem with the applicability of TE, ETT, and counterfactual fairness is the computation of the non-observable terms in Eqs~\ref{eq:TE},~\ref{eq:ETT}, and~\ref{eq:counterfactual}. These terms are either interventional (e.g. $P(\hat{y}_{A\leftarrow a_1}$)) or counterfactual (e.g. $P(\hat{Y}_{A\leftarrow a_1}=\hat{y} \mid X=x, A = a_0)$. In scenarios where these quantities can be expressed in terms of observable probabilities (e.g. joint probabilities, conditional probabilities, etc.), it is said that they are \textit{identifiable}. Otherwise, they are unidentifiable. Typically, the identifiability of interventional and counterfactual quantities depends on the structure of the causal graph~\citep{shpitser08,pearl2009causality}. Alternatively, if  all parameters of the causal model are known (including the latent variables distributions $P(U=u)$), any counterfactual is identifiable and can be computed using the three steps abduction, action, and prediction (Theorem 7.1.7 in~\citep{pearl2009causality}). The details of the computation of a counterfactual probability using a simple deterministic example are provided in~\ref{ap:abductionComputation}.

A simple but important implication of Eq.~\ref{eq:counterfactual} is that, given a causal graph, a predictor $\hat{Y}$ is counterfactually fair if it is a function of non-descendants of the sensitive variable $A$. In other words, if $\hat{Y}$ is a function of variables that depend on $A$ (there is a directed path between any one of those variables and $A$), it is not counterfactually fair. Consequently, one can tell if a predictor is counterfactually fair by simply checking the causal graph\footnote{Kusner et al.~\citep{kusner2017counterfactual} identify some exceptions, but guaranteeing that they will \textit{not happen in general}.}.


No unresolved discrimination~\citep{kilbertus2017avoiding} is another causal-based fairness notion which is satisfied when no directed paths from the sensitive attribute $A$ to the predictor $\hat{Y}$ are allowed, except via a resolving variable. A resolving variable is any variable in a causal graph that is influenced by the sensitive attribute in a manner that is accepted as nondiscriminatory (this is similar to explanatory attributes in conditional statistical parity (Section~\ref{sec:notioncsp})). In the job hiring example, if we assume that the effect of $A$ on the education level is nondiscriminatory,
it implies that the differences in education level for different values of $A$ are not considered as discrimination. Thus, a disparity in the predictions between men and women might been explained and justified by their corresponding education levels. Hence, the education level acts as a resolving variable. 
Figure~\ref{fig:unresolved} shows two similar causal graphs for our hiring example, yet differ in some of the causal relations between variables. By considering the education as a resolving variable, the graph at the left exhibits unresolved discrimination along the dashed paths: $A \rightarrow Experience \rightarrow \hat{Y}$ and $A \rightarrow \hat{Y}$. By contrast, the graph at the right does not exhibit any unresolved discrimination as the effect of $A$ on $\hat{Y}$ is justified by the resolved variable Education: $A \rightarrow Education \rightarrow \hat{Y}$. 

\begin{figure}[!h]
	\centering
	\subfloat{{\includegraphics[scale=0.25]{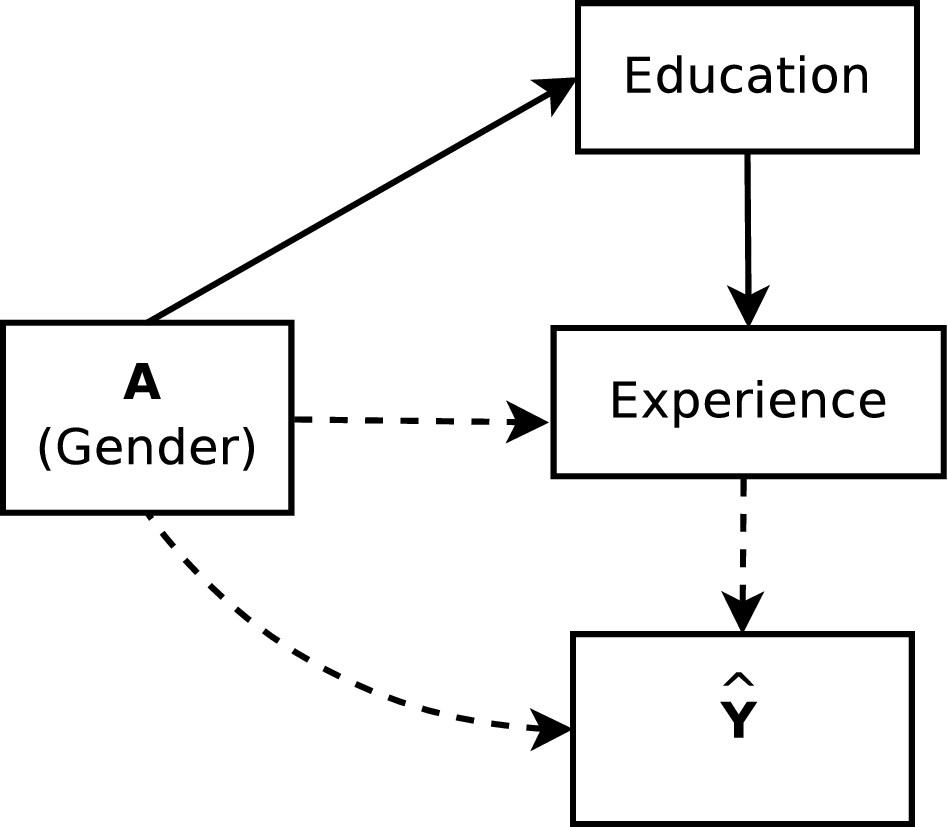}}}
	\qquad \qquad
	\subfloat{{\includegraphics[scale=0.25]{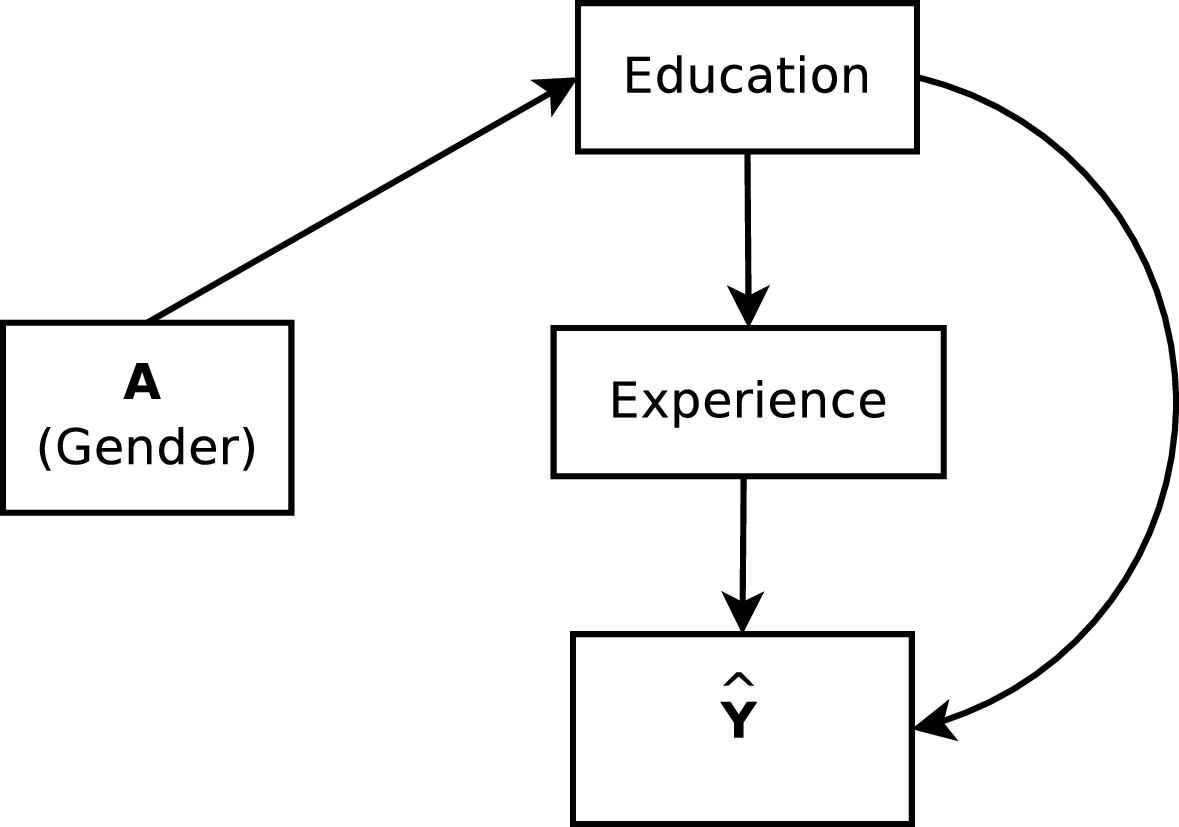}}}
	\caption{Two possible graphs for the hiring example. If \textit{Education} is a resolving variable, the predictor $\hat{Y}$ exhibits unresolved discrimination in the left graph (along the dashed paths), but not in the right one.}
	\label{fig:unresolved}
\end{figure}

No unresolved discrimination is equivalent to other fairness notions in some interesting special cases~\citep{kilbertus2017avoiding}. For instance, if no resolving variables exist, no unresolved discrimination is analogous to statistical parity (Section~\ref{sec:notion_sp}) in a causal context. $A$ and $\hat{Y}$ are statistically independent and no directed paths from $A$ to $\hat{Y}$ are allowed. Likewise, no unresolved discrimination might be equivalent to equalized odds (Section~\ref{sec:eqOdds}) in a causal context if the set of resolving variables is the singleton set of actual outcomes: $\{Y\}$. Compared to counterfactual fairness, no unresolved discrimination is a weaker notion. That is, a counterfactually unfair scenario may be identified as fair based on no unresolved discrimination. This can happen in case one or several variables in the causal graph are identified as resolving.

A causal graph exhibits potential proxy discrimination~\citep{kilbertus2017avoiding} if there exists a path from the protected attribute $A$ to the predicted outcome $\hat{Y}$ that is blocked by a proxy variable $P_x$. A proxy is a descendant of $A$ that is chosen to be labelled as a proxy because it is significantly correlated with $A$. Given a causal graph, a predictor $\hat{Y}$ exhibits no proxy discrimination if following equality holds for all potential proxies $P_x$. 
\begin{equation}
\label{eq:proxy}
P(\hat{Y}_{P_x \leftarrow p}) = P(\hat{Y}_{P_x \leftarrow p\prime}) \quad \forall \; p, p\prime
\end{equation}
In other words, Eq.~\ref{eq:proxy} implies that changing the value of $P_x$ should not have any impact on the prediction. 

In the job hiring example, the job experience can be considered as a proxy of an individual's gender. Figure~\ref{fig:proxy} shows two similar causal graphs. The one at the left presents a potential proxy discrimination via the path: $A \rightarrow Experience \rightarrow \hat{Y}$. However, the graph at the right is free of proxy discrimination as the edge between $A$ and its proxy $P_x$ (here Experience) has been removed along with all incoming arrows of $P_x$ (the edge between \textit{Education} and \textit{Experience}). 

\begin{figure}[!h]
    \centering
    \subfloat {{\includegraphics [scale=0.25]{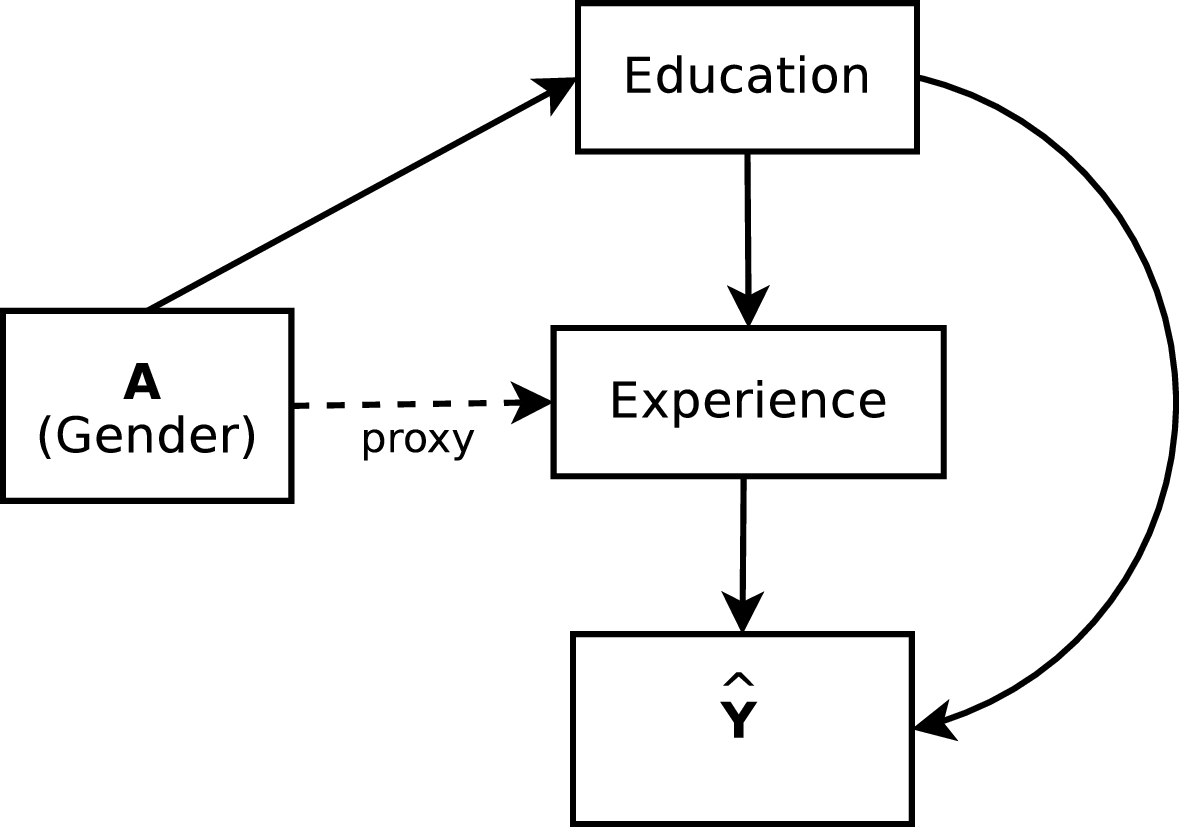} }}
    \qquad \qquad
    \subfloat {{\includegraphics[scale=0.25]{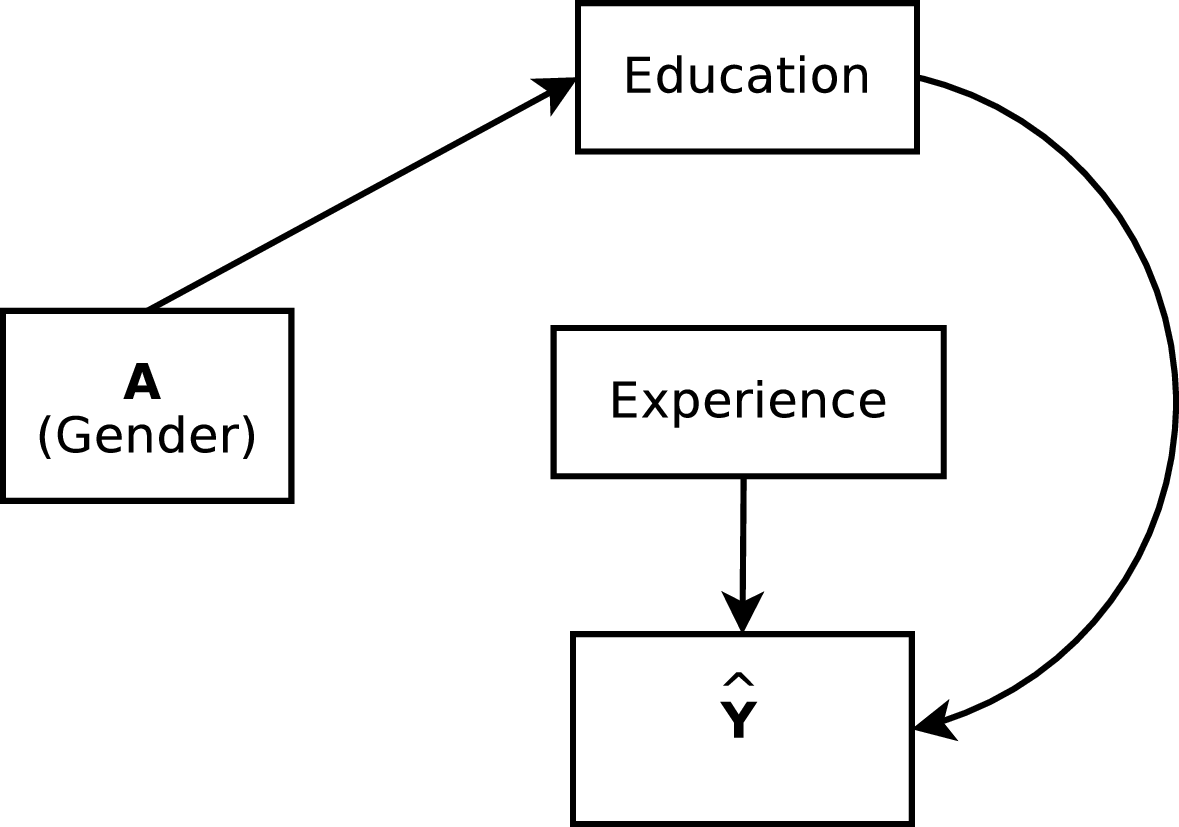} }}
    \caption{Two possible graphs to describe proxy discrimination. If we consider \textit{Experience} as a proxy of the sensitive attribute A, the graph at the left exhibits a potential proxy discrimination (along the dashed edge between A and Experience), but not in the right one.}
    \label{fig:proxy}
\end{figure}

Other causal based fairness notions include direct/indirect effect~\citep{pearl01direct}, FACE/FACT~\citep{khademi2019fairness}, counterfactual effects~\citep{zhang2018fairness}, counterfactual error rates~\citep{zhang2018equality}, and path-specific counterfactual fairness~\citep{chiappa2019path,wu2019pc}.

As a general rule, causality-based fairness notions can be used as long as the causal relationships between the attributes are identified and represented using a reliable and plausible causal graph. The construction of the causal graph requires typically domain-specific expertise and can be validated by existing datasets. In practice, however, causality-based fairness notions are recommended in at least two notable scenarios. The first scenario is when the legal framework of the case at hand is disparate treatment. In such framework, to win a discrimination case, the plaintiff must show that the defendant has used (directly or indirectly (via proxy)) the sensitive attribute $A$ to take the discriminatory decision $\hat{Y}$. In other words, she must prove that the variable $A$ is a cause of $\hat{Y}$ \textcolor{black}{while the causal effect of $A$ on $\hat{Y}$ is central to all causal-based fairness notions mentioned above}. The second scenario is when there is confounding between $A$ and $\hat{Y}$. That is, there is a covariate which is a common cause of $A$ and $\hat{Y}$. Such scenario can lead to statistical anomalies such as Simpson's paradox~\citep{simpson1951interpretation,pearl2009causality} where the statistical conclusions drawn from the sub-populations differ from that from the whole population. The Berkeley admission case~\citep{berkeley75} is a known real-world example of such statistical anomaly. In such scenarios, any statistical fairness notion which relies solely on correlation between variables, will fail to detect bias. Hence, causality-based fairness notions are necessary to appropriately address the problem of fairness.   
}

\section{Relaxation}
\label{sec:relaxation}
Almost all fairness notions presented so far involve a strict equality between quantities, in particular probabilities. In real scenarios, however, it is more suitable to opt for an approximate or relaxed form of fairness constraint. The need for relaxation might be due to the impossibility to apply fairness strictly on the application at hand, or merely, it is not a requirement to impose an exact constraint~\citep{kim2020model}.

Fairness notion definitions can be relaxed by considering a threshold on the ratio or difference between quantities. For instance, the requirement for statistical parity (Section~\ref{sec:notion_sp}) can be relaxed in one of the two following ways:
\begin{itemize}
\item By allowing the ratio between the predicted acceptance rates of protected and unprotected groups to reach the threshold of $\epsilon$ (a.k.a $p\%$ rule defined as satisfying this inequality when $\epsilon = p/100$ \citep{zafar2015fairness}): 
\begin{equation}
	\label{eq:relRatio}
\frac{P(\hat{Y} \mid A = 0)}{P(\hat{Y} \mid A = 1)} \geq 1-\epsilon \quad \forall \; \epsilon \in [0,1]
\end{equation}
For $\epsilon = 0.2$, this condition relates to the 80\% rule in disparate impact law \citep{feldman2015certifying,barocas2016big}.
\item By allowing the difference between the predicted acceptance rates of different groups to reach a threshold of $\epsilon$~\citep{dwork2012fairness}:
\begin{equation}
\label{eq:relDiff}
\mid P(\hat{Y} \mid A = 0) - P(\hat{Y} \mid A = 1) \mid \; \leq \epsilon \quad \forall \; \epsilon \in [0,1]
\end{equation}
\end{itemize}

A notable difference between the two types of relaxation is that the second one (Eq.~\ref{eq:relDiff}) is insensitive to which group/individual is the victim of discrimination as the formula is using absolute value.

Fairness through awareness can be relaxed using three threshold values, $\alpha_1, \alpha_2,$ and $\gamma$ as follows~\citep{yona2018probably}:

\begin{equation}
\label{eq:relFTA}
P\biggl[ P\Bigl[ \bigl|M(v_i)-M(v_j)\bigr| > d(v_i, v_j) + \gamma\;\; \Bigr]   >\alpha_2\biggr] \leq \alpha_1.
\end{equation} 

The relaxation is allowing $M(v_i)-M(v_j)$ to exceed $d(v_i,v_j)$ by a margin of $\gamma$, but the fraction of individuals differing from them by $\gamma$ should not exceed $\alpha_2$. If the fraction exceeds $\alpha_2$, the individual is said to be $\alpha_2$-discriminated against.

To allow for more flexibility in the application of fairness notions, other relaxations can be considered. For instance, Eq.~\ref{eq:csp} of conditional statistical parity (Section~\ref{sec:notioncsp}) can be modified by relaxing the strict equality $E=e$ as follows:

\begin{equation}
\label{eq:relcsp}
P(\hat{Y}=1 \mid e-\epsilon \leq E \leq e+\epsilon, A = 0) = P(\hat{Y}=1 \mid e-\epsilon \leq E \leq e+\epsilon, A = 1) 
\end{equation}

\section{Classification and tensions}
\label{sec:trade-offs}

Group fairness notions fall into three classes defined in terms of the properties of joint distributions, namely, independence, separation, and sufficiency~\citep{barocas-hardt-narayanan}. These properties are used in the literature to prove the existing of tensions between fairness notions, that is, it is impossible to satisfy all fairness notions simultaneously except in extreme, degenerate, and dump scenarios. Besides, the applicability of most of fairness notions can be ameliorated by relaxing their strict definitions.

\subsection{Classification}
\label{sec:classif}
Group fairness (a.k.a statistical fairness) notions can be characterized by the properties of the joint  distribution of the sensitive attribute $A$, the label $Y$, and the classifier $\hat{Y}$ (or score $S$). This means that we can write them as some statement involving properties of these three random variables resulting in the three following fairness criteria~\citep{barocas2017fairness, barocas-hardt-narayanan}:

\paragraph{Independence} Independence means that the sensitive feature $A$ is statistically independent of the classifier $\hat{Y}$ (or the score $S$).
\begin{equation}
\label{eq:ind}
\hat{Y}  \perp A \quad (or \; S \perp A)
\end{equation}
In the case of binary classification, independence is equivalent to statistical parity as defined in Section~\ref{sec:notion_sp},  Eq.~\ref{eq:sp}. Conditioning on \textcolor{black}{explanatory} variables ($E$) yields a variant of independence as follows.

\paragraph{Conditional independence} 

\begin{equation}
\label{eq:ind}
\hat{Y}  \perp A \mid E \quad (or \; S \perp A \mid E)
\end{equation}
This class includes conditional statistical parity defined in Section~\ref{sec:notioncsp}, Eq.~\ref{eq:csp}.

\paragraph{Separation} Separation denotes a class of fairness notions satisfying, at different degrees, conditional independence between the prediction $\hat{Y}$ and the sensitive attribute $A$ given the actual outcome $Y$. 
\begin{equation}
\label{eq:sep}
\hat{Y} \perp A \mid Y \quad (or \; S \perp A \mid Y)
\end{equation}

In the case where $\hat{Y}$ is a binary classifier, the formulation of separation is equivalent to that of the equalized odds (Eq.~\ref{eq:eqOdds}). Equal opportunity (Eq.~\ref{eq:eqOpp}), predictive equality (Eq.~\ref{eq:predEq}), balance for positive class (Eq.~\ref{eq:balPosclass}), and balance for negative class (Eq.~\ref{eq:balNegclass}) are all relaxations of separation. Some incompatibility results do hold for separation, but do not hold for the relaxations. More on this in the next section~(Section~\ref{sec:tension}).

\paragraph{Sufficiency} Sufficiency is a class of fairness notions satisfying, at different degrees, conditional independence between the target variable $Y$ and the sensitive attribute $A$ given the prediction $\hat{Y}$. 
\begin{equation}
\label{eq:suff}
Y \perp A \mid \hat{Y} \quad (or \;Y \perp A \mid S)
\end{equation}
In the case of binary classification, strict sufficieny corresponds to conditional use accuracy equality (Eq.~\ref{eq:condUseAcc}). Using the score $S$, calibration (Eq.~\ref{eq:calib}), and well-calibration (Eq.~\ref{eq:wellCalib}) can be considered as sufficiency~\citep{chouldechova2017fair}. Relaxation of sufficiency yields to predictive parity (Eq.~\ref{eq:predPar}) which also does not satisfy exactly the same incompatibility result as sufficiency (Section~\ref{sec:tension}). 

Table~\ref{tab:summary} lists all fairness notions along with their classification.

\begin{landscape}
\begin{table}
\setstretch{1.0}
 \caption{Classification of fairness notions. ($*$ notion newly defined in this paper)}
 \label{tab:summary}
 \setlength\extrarowheight{7pt} 
\begin{tabular}{|l|l|l|c|l|}
\hline 
\footnotesize{Fairness Notion} & \footnotesize{Ref.}& \footnotesize{Formulation} & \footnotesize{Classification} & \footnotesize{Type} \\
\hline
 \footnotesize{Statistical parity}   &  \footnotesize{\citep{dwork2012fairness}} & \footnotesize{$P(\hat{Y} \mid A = 0) = P(\hat{Y} \mid A = 1) $}    & \makecell{\footnotesize{Independence} \\ \footnotesize{(equivalent or relaxed$^\bigstar$)}} &   \\
  \cline{1-3}
  \footnotesize{Conditional statistical parity} &\footnotesize{\citep{corbett2017algorithmic}}  &  \footnotesize{$P(\hat{Y}=1 \mid E=e,A = 0) = P(\hat{Y}=1 \mid E=e,A = 1)$}$^\bigstar$   &   &  \\
\cline{1-4}
     \footnotesize{Equalized odds} &{\multirow{2}{*}{\footnotesize{\citep{hardt2016equality}}}} & {\footnotesize $P(\hat{Y} = 1 \mid Y=y,\; A=0) = P(\hat{Y}=1 \mid Y= y,\; A=1)  \quad \forall{ y \in \{0,1\}}$} &  & \\
   \cline{1-1}  \cline{3-3}
  \footnotesize{Equal opportunity} &      &   \footnotesize{$P(\hat{Y}=1 \mid Y=1,A = 0) = P(\hat{Y}=1\mid Y=1,A = 1)$}$^\bigstar$    & \makecell{\footnotesize{Separation} \\ \footnotesize{(equivalent or relaxed$^\bigstar$)}} & \\
    \cline{1-3}
   \footnotesize{Predictive equality} & \footnotesize{\citep{corbett2017algorithmic}}& \footnotesize{$P(\hat{Y}=1 \mid Y=0,A = 0) = P(\hat{Y}=1\mid Y=0,A = 1)$}$^\bigstar$  &  & \\
   \cline{1-3}
   \footnotesize{Balance for positive class}  & {\multirow{2}{*}{\footnotesize{\citep{kleinberg_et_al:LIPIcs:2017:8156}}}}& \footnotesize{$E[S \mid Y =1,A = 0)] = E[S \mid Y =1,A = 1]^\bigstar$}  &   &  \\
 \cline{1-1}  \cline{3-3}
        \footnotesize{Balance for negative class} &  & \footnotesize{$E[S \mid Y =0,A = 0] = E[S \mid Y =0,A = 1]^\bigstar$} & & \\
 \cline{1-3}
        \footnotesize{Overall balance} & \makecell{\footnotesize{*}} & \footnotesize{$E[S \mid Y =y,A = 0] = E[S \mid Y =y,A = 1] \quad \forall y \in \{0,1\}$} & & \makecell{\parbox[c]{-2mm}{\multirow{5}{*}{\rotatebox[origin=c]{-90}{\footnotesize {Group}}}}}\\
 \cline{1-4}
\footnotesize{Conditional use acc. equality} &\footnotesize{\citep{berk2018fairness}} & {\footnotesize $P(Y=y\mid \hat{Y}=y ,A = 0) = P(Y=y\mid \hat{Y}=y,A = 1) \quad \forall{ y \in \{0,1\}}$} &      &  \\
     \cline{1-3}
    \footnotesize{Predictive parity} & \footnotesize{\citep{chouldechova2017fair}}& \footnotesize{$P(Y=1 \mid \hat{Y} =1,A = 0) = P(Y=1\mid \hat{Y} =1,A = 1)^\bigstar$}  &.   \makecell{\footnotesize{Sufficiency} \\ \footnotesize{(equivalent or relaxed$^\bigstar$)}} & \\
 \cline{1-3} 
 \footnotesize{Negative predictive parity} &\makecell{\footnotesize{*}} &  \footnotesize {$ P(Y=1 \mid \hat{Y} =0,A = 0) = P(Y=1\mid \hat{Y} =0,A = 1)^\bigstar$}  & & \\
 \cline{1-3}  
 \footnotesize{Calibration} & \small{\citep{chouldechova2017fair}} &  \footnotesize {$P(Y =1 \mid S =s,A = 0) = P(Y =1 \mid S =s,A = 1) \quad \forall{ s \in [0,1]} $} &  &\\
  \cline{1-3}
 \footnotesize{Well-calibration}  & \footnotesize{\citep{kleinberg_et_al:LIPIcs:2017:8156}}& \footnotesize {$P(Y =1 \mid S =s,A = 0) = P(Y =1 \mid S =s,A = 1) = s  \quad  \forall \; {s \in [0,1]}$}  &   &\\
   \cline{1-4}
\footnotesize{Overall accuracy equality}    &   &  \footnotesize{$ P(\hat{Y} = Y | A = 0) = P(\hat{Y} = Y | A = 1)$}     & \makecell{\footnotesize{Other metrics} }   & \\     
 \cline{1-1}  \cline{3-3}
\textcolor{black}{\footnotesize{Treatment equality}} &{\multirow{3}{*}{\footnotesize{\citep{berk2018fairness}}}} & \textcolor{black}{\makecell [l] {$\frac{FN}{FP} \textsubscript{(A=0)} = \frac {FN}{FP} \textsubscript{(A=1)}$}} & \makecell{\footnotesize {from confusion matrix}\\\\} & \\
 \cline{1-1}  \cline{3-4}
\footnotesize{Total fairness} & & \makecell{$-$} & \makecell{ \footnotesize {Independence, Separation}\\ \footnotesize{and Sufficiency}}  & \\
\cline{1-4}
\textcolor{black}{\footnotesize{Total effect}} &\textcolor{black}{{\multirow{2}{*}{\footnotesize{\citep{pearl2009causality}}}}} & \textcolor{black}{\footnotesize{$TE_{a_1,a_0} (\hat{y}) = P(\hat{y}_{A\leftarrow a_1}) - P(\hat{y}_{A\leftarrow a_0})$}} & & \\
 \cline{1-1}  \cline{3-3}
\textcolor{black}{\footnotesize{Effect of treatment on treated}} & & \textcolor{black}{\footnotesize{$ETT_{a_1,a_0} (\hat{y}) = P (\hat{y}_{A\leftarrow a_1} \mid a_0) - P (\hat{y} \mid a_0)$}}& & \\
    \cline{1-3}
           \footnotesize{No unresolved discrimination} &{\multirow{2}{*}{\footnotesize {\citep{kilbertus2017avoiding}}}} &  \makecell{$-$} & \makecell{\footnotesize {Causality}}& \\
 \cline{1-1}  \cline{3-3}
 \footnotesize{No proxy discrimination} &  & \footnotesize{$P(\hat{Y} \mid do(P_x=p)) = P(\hat{Y} \mid do(P_x=p\prime)) \quad \forall P_x \quad and \quad \forall \; p, p\prime$} & & \\
\cline{1-3} \cline{5-5}
       \footnotesize{Counterfactual fairness} & \footnotesize{\citep{kusner2017counterfactual}} & {\footnotesize $P(\hat{Y}_{A \leftarrow a} (U)  =y  \mid X =  x, A =  a) = P(\hat{Y}_{A \leftarrow a\textprime} (U)  =y  \mid X =  x, A =  a) $} & & \makecell{\parbox[c]{-2mm}{\multirow{3}{*}{\rotatebox[origin=c]{-90}{\footnotesize {Individual}}}}} \\
    \cline{1-4}
   \textcolor{black}{\footnotesize{Causal discrimination}} &  \footnotesize{\citep{galhotra2017fairness}} & \textcolor{black}{\footnotesize{$X\textsubscript{(A=0)} = X \textsubscript{(A=1)}  \; \land \; A\textsubscript{(A=0)} \; \neq  A\textsubscript{(A=1)} \; \Rightarrow \hat{y}\textsubscript{(A=0)} =  \hat{y}\textsubscript{(A=1)}$}} & \makecell{{\multirow{2}{*}{\footnotesize {Similarity Metric}}}} & \\
\cline{1-3}
       \footnotesize{Fairness through awareness}  & \footnotesize{\citep{dwork2012fairness}}  & \footnotesize{$D(M(v_i), M(v_j))  \leq d(v_i, v_j)$} &  &\\
   \hline
\end{tabular}
\end{table}
\end{landscape}

\subsection{Tensions}
\label{sec:tension}

It has been proved that there are incompatibilities between fairness notions. That is, it is not always possible for an MLDM to satisfy specific fairness notions simultaneously~\citep{barocas2017fairness,barocas-hardt-narayanan,chouldechova2017fair,zafar2017fairness,mitchell2018prediction}. In presence of such incompatibilities, the MLDM should make a trade-off to satisfy some notions on the expense of others or partially satisfy all of them. Incompatibility\footnote{The term impossibility is commonly used as well.} results are well summarized by Mitchell et al.~\citep{mitchell2018prediction} as follows:

\paragraph{Statistical parity (independence) versus conditional use accuracy equality (sufficiency)} Independence and sufficiency are incompatible, except when both groups (protected and non-protected) have equal base rates or $\hat{Y}$ and $Y$ are independent. Note, however, that $\hat{Y}$ and $Y$ should not be independent since otherwise the predictor is completely useless.  More formally,   
\begin{center}
\begin{tabular}{ccccccc}
	$\hat{Y}  \perp A$ &  AND & $Y \perp A \mid \hat{Y}$  &  $\Rightarrow$ &  $Y \perp A$ & OR & $\hat{Y} \perp Y$ \\
	(independence) &   & (strict sufficiency) &  & (equal base rates) & & (useless predictor) \\
\end{tabular}
\end{center}

It is important to mention here that this result does not hold for the relaxation of sufficiency, in particular, predictive parity. Hence, it is possible for the output of an MLDM to satisfy statistical parity and predictive parity between two groups having different base rates. Such example needs to satisfy the following constraints, assuming two groups $a$ and $b$:

\begin{center}
	\renewcommand{\arraystretch}{2}
\begin{tabular}{ccclc}
	&  & $ \frac{TP_a + FP_a}{TP_a + FP_a + FN_a + TN_a} = \frac{TP_b + FP_b}{TP_b + FP_b + FN_b + TN_b}$ & (independence) \\
	&  & $ \frac{TP_a}{TP_a + FP_a} = \frac{TP_b}{TP_b + FP_b}$ & (predictive parity) \\
	&  & $ \frac{TP_a + FN_a}{TP_a + FP_a + FN_a + TN_a} \neq \frac{TP_b + FN_b}{TP_b + FP_b + FN_b + TN_b}$ & (different base rates) \\
\end{tabular}

\renewcommand{\arraystretch}{1}
\end{center}

An example scenario satisfying the above constrains is the following:

\begin{center}
\renewcommand{\arraystretch}{1.5}
$\begin{array}{c}
	PPV_a = 0.4 \\ 
	baserate_a = 0.43 \\
\end{array}$
$\begin{array}{c|c}
TP_a = 9 & FP_a = 6  \\ \hline
FN_a = 4 & TN_a = 11 \\
\end{array}$
$\qquad$
$\begin{array}{c|c}
TP_b = 12 & FP_b = 8  \\ \hline
FN_b = 2 & TN_b = 18 \\ 
\end{array}$
$\begin{array}{c}
	PPV_b = 0.4 \\ 
	baserate_b = 0.35 \\
\end{array}$
\renewcommand{\arraystretch}{1}
\end{center}

\paragraph{Statistical parity (independence) versus equalized odds (separation)} Similar to the previous result, independence and separation are mutually exclusive unless base rates are equal or the predictor $\hat{Y}$ is independent from the actual label $Y$~\citep{barocas-hardt-narayanan}. As mentioned earlier, dependence between $\hat{Y}$ and $Y$ is a weak assumption as any useful predictor should satisfy it. More formally, 
\begin{center}
	\begin{tabular}{ccccccc}
	$\hat{Y}  \perp A$ &  AND & $\hat{Y} \perp A \mid Y$  & $\Rightarrow$ &  $Y \perp A$  & OR  & $\hat{Y} \perp Y$ \\
	(independence) &   & (strict separation) &  & (equal base rates)  &  &  (useless predictor)
\end{tabular}
\end{center}

Considering a relaxation of equalized odds, that is, equal opportunity or predictive equality, breaks the incompatibility between independence and separation. An MLDM whose output satisfies independence and equal opportunity, but with different base rates between groups should satisfy the following constraints:

\begin{center}
	\renewcommand{\arraystretch}{2}
\begin{tabular}{ccclc}
	&  & $ \frac{TP_a + FP_a}{TP_a + FP_a + FN_a + TN_a} = \frac{TP_b + FP_b}{TP_b + FP_b + FN_b + TN_b}$ & (independence) \\
	&  & $\frac{TP_a}{TP_a + FN_a} = \frac{TP_b}{TP_b + FN_b}$ & (equal opportunity) \\
	&  & $\frac{TP_a + FN_a}{TP_a + FP_a + FN_a + TN_a} \neq \frac{TP_b + FN_b}{TP_b + FP_b + FN_b + TN_b}$ & (different base rates) \\
\end{tabular}

\renewcommand{\arraystretch}{1}
\end{center}

An example scenario satisfying the above constrains is the following:

\begin{center}
\renewcommand{\arraystretch}{1.5}
$\begin{array}{c}
	TPR_a = 0.6 \\ 
	baserate_a = 0.55 \\
\end{array}$
$\begin{array}{c|c}
TP_a = 9 & FP_a = 3  \\ \hline
FN_a = 2 & TN_a = 6 \\
\end{array}$
$\qquad$
$\begin{array}{c|c}
TP_b = 12 & FP_b = 6  \\ \hline
FN_b = 8 & TN_b = 4 \\ 
\end{array}$
$\begin{array}{c}
	TPR_b = 0.6 \\ 
	baserate_b = 0.71 \\
\end{array}$
\renewcommand{\arraystretch}{1}
\end{center}

\paragraph{Equalized odds (separation) vs conditional use accuracy equality (sufficiency)} Separation and sufficiency are mutually exclusive, except in the case where groups have equal base rates. More formally:
\begin{center}
	\begin{tabular}{ccccccc}
		$\hat{Y} \perp A \mid Y$ &  AND & $Y \perp A \mid \hat{Y}$  & $ \Rightarrow$ &  $Y \perp A$  & & \\
	(strict separation) &   & (strict sufficiency) &  & (equal base rates)  &  &  \\
\end{tabular}
\end{center}

Both separation and sufficiency have relaxations. Considering only one relaxation will only drop the incompatibility for extreme and degenerate cases. For example, predictive parity (relaxed version of sufficiency) is still incompatible with separation (equalized odds), except in the following three extreme cases~\citep{chouldechova2017fair}:
\begin{itemize}
\item both groups have equal base rates.
\item both groups have $FPR=0$ and $PPV=1$.
\item both groups have $FPR=0$ and $FNR=1$.
\end{itemize}

The incompatibility disappears completely when considering relaxed versions of both separation and sufficiency. For example, the following scenario satisfies equal opportunity (relaxed version of separation) and predictive parity (relaxed version of sufficiency) while base rates are different in both groups:
\begin{center}
\renewcommand{\arraystretch}{1.5}
$\begin{array}{c}
	TPR_a = 0.4 \\ 
	PPV_a = 0.75 \\ 
	baserate_a = 0.6 \\
\end{array}$
$\begin{array}{c|c}
TP_a = 9 & FP_a = 6  \\ \hline
FN_a = 3 & TN_a = 2 \\
\end{array}$
$\qquad$
$\begin{array}{c|c}
TP_b = 12 & FP_b = 8  \\ \hline
FN_b = 4 & TN_b = 8 \\ 
\end{array}$
$\begin{array}{c}
	TPR_b = 0.4 \\ 
	PPV_b = 0.75 \\ 
	baserate_b = 0.5 \\
\end{array}$
\renewcommand{\arraystretch}{1}
\end{center}

\subsection{Group vs individual fairness}
\label{sec:gfvsif}
Compared to individual fairness notions, the main concern for group fairness notions is that they are only suited to a limited number of coarse-grained, predetermined protected groups based on some sensitive attribute (e.g. gender, race, etc.). Hence group fairness notions are not suitable in presence of intersectionality~\citep{crenshaw1990mapping} where individuals are often disadvantaged by multiple sources of discrimination: their race, class, gender, religion, and other inner traits. Typically, statistical fairness can only be applied across a small number of coarsely defined groups, and hence failing to identify discrimination on structured subgroups (e.g. single women) known also as ``fairness gerrymandering''~\citep{kearns2018preventing}. A simple alternative might be to apply statistical fairness across every possible combination of protected attributes. There are at least two problems to this approach. First, this can lead to an impossible statistical problem with the large number of sub-groups which may lead in turn to overfitting. Second, groups which are not (yet) defined in anti-discrimination law may exist and may need protection~\citep{wachter2019right}. Another issue with group fairness notions is their susceptibility to masking. Most of group fairness notions can be gamed by adding arbitrarily selected samples to satisfy the fairness notion formula, that is, to just ``make up the numbers''.

Compared to group fairness notions, individual fairness notions have the drawback that they can result in ``unjust disparities in outcomes between groups''~\citep{binns2020apparent}. For illustration, consider the example in Table~\ref{tab:groupIndIssue} where fairness through awareness is satisfied (Eq.~\ref{eq_FTA}) whereas statistical parity Eq.~(\ref{eq:sp}) is not. Fairness through awareness is satisfied since for every pair of candidates, the distance between the probability distributions on the outcomes ($M()$) is smaller than the distance between the pair of candidates. On the other hand, if the hiring threshold is $0.6$, only one female candidate ($F2$) will be hired as she has a probability of acceptance $P(\hat{Y}=1) = 0.8 > 0.6$ whereas all male candidates will be hired. Another important issue for similarity-based individual fairness (e.g. fairness through awareness) is the difficulty to obtain a similarity value between every pair of individuals. For example, even with the assumption that the similarity can be quantified between all individuals in the training data, it might be challenging to generalize to new individuals~\citep{binns2020apparent}.

\begin{table}[!h]
\caption{A job hiring scenario satisfying fairness through awareness (Eq.~\ref{eq_FTA}) but not statistical parity (Eq.~\ref{eq:sp}) for a threshold of $0.6$.  The second row ($M()$) indicates the probability distribution on the outcomes. For example, for the first female applicant $F1$, $P(\hat{Y}=1)=0.58$ and $P(\hat{Y}=0)=0.42$. Each cell at the left of the shaded table's diagonal represents a distance between a pair of applicants. Those at the right represent the distance between probability distributions on the outcomes. 
 \medskip}
\centering
\label{tab:groupIndIssue}    
  \begin{tabular}{|c|c|c|c|c|c|c|l}
\cline{1-7}
 & F1 & F2 & F3 & M1 & M2 & M3 & \\ \cline{1-7}
   $\mathbf{M()}$            & $[0.58,0.42]$ & $[0.8,0.2]$ & $[0.55,0.45]$ &  $[0.65,0.35]$ & $[0.81,0.19]$ & $[0.61,0.39]$ & \\ \cline{1-7} \cline{1-7} 
 F1            & {\cellcolor{gray}} & 0.17 & 0.021 & 0.051 & 0.18 & 0.02 & {\small{\multirow{5}{*}{\rotatebox[origin=c]{90}{$D(M(v_i),M(v_j))$}}}} \\
\cline{1-7} 
    F2            & 0.21 & {\cellcolor{gray}}& 0.19& 0.11 & 0.008 & 0.15\\
\cline{1-7}
    F3             & 0.06  & 0.22&{\cellcolor{gray}} &0.07 & 0.20 & 0.04\\
\cline{1-7}
    M1             & 0.1 & 0.15&0.1 & {\cellcolor{gray}}&0.12 & 0.029\\
\cline{1-7}
    M2        &  0.2 & 0.01 & 0.3 &0.15 &{\cellcolor{gray}} & 0.15\\
\cline{1-7}
   M3            &0.05  & 0.17 & 0.08 &0.05 & 0.17& {\cellcolor{gray}} \\ \cline{1-7} 
\multicolumn{1}{l}{}&\multicolumn{1}{l}{} & \multicolumn{1}{l}{}& \multicolumn{1}{l}{$d(v_i,v_j)$}&\multicolumn{1}{l}{} & \multicolumn{1}{l}{}&\multicolumn{1}{l}{} \\
\end{tabular}
\end{table}

Several researchers assume that both group and individual fairness are prominent, yet, conflicting and suggest approaches to minimize the trade-offs between these notions \citep{binns2020apparent}. For instance, \citep{friedler2016possibility} define two different worldviews, WYSIWYG and WAE. The WYSIWYG (What you see is what you get) worldview assumes that the unobserved (construct) space and observed space are essentially the same while the WAE (we're all equal) worldview implies that there are no innate differences between groups of individuals based on certain potentially discriminatory characteristics. These two worldviews highlight the tension between group and individual fairness. For instance, in the job hiring example, the WYSIWYG might be the assumption that attributes like education level and job experience (which belong to the observed space) correlate well with the applicant's seriousness or hardworking (properties of the construct space). This is to say that there is some way to combine these two spaces to correctly compare true applicant aptitude for the job. On the other hand, the WAE claims that all groups will have almost the same distribution
in the construct space of inherent abilities (here, seriousness and hardworking), chosen as important inputs to the decision making process. The idea is that any difference in the groups’ performance (e.g., academic achievement or education level) is due to factors outside their individual control (e.g., the quality of their neighborhood school) and should not be taken into account in the decision making process. Thus, the choice between fairness notions must be based on an explicit choice in worldviews. 

\section{Diagram and discussion}
\label{sec:diagram}

\begin{landscape}
\begin{figure}[ht]
    \includegraphics [scale=0.25]{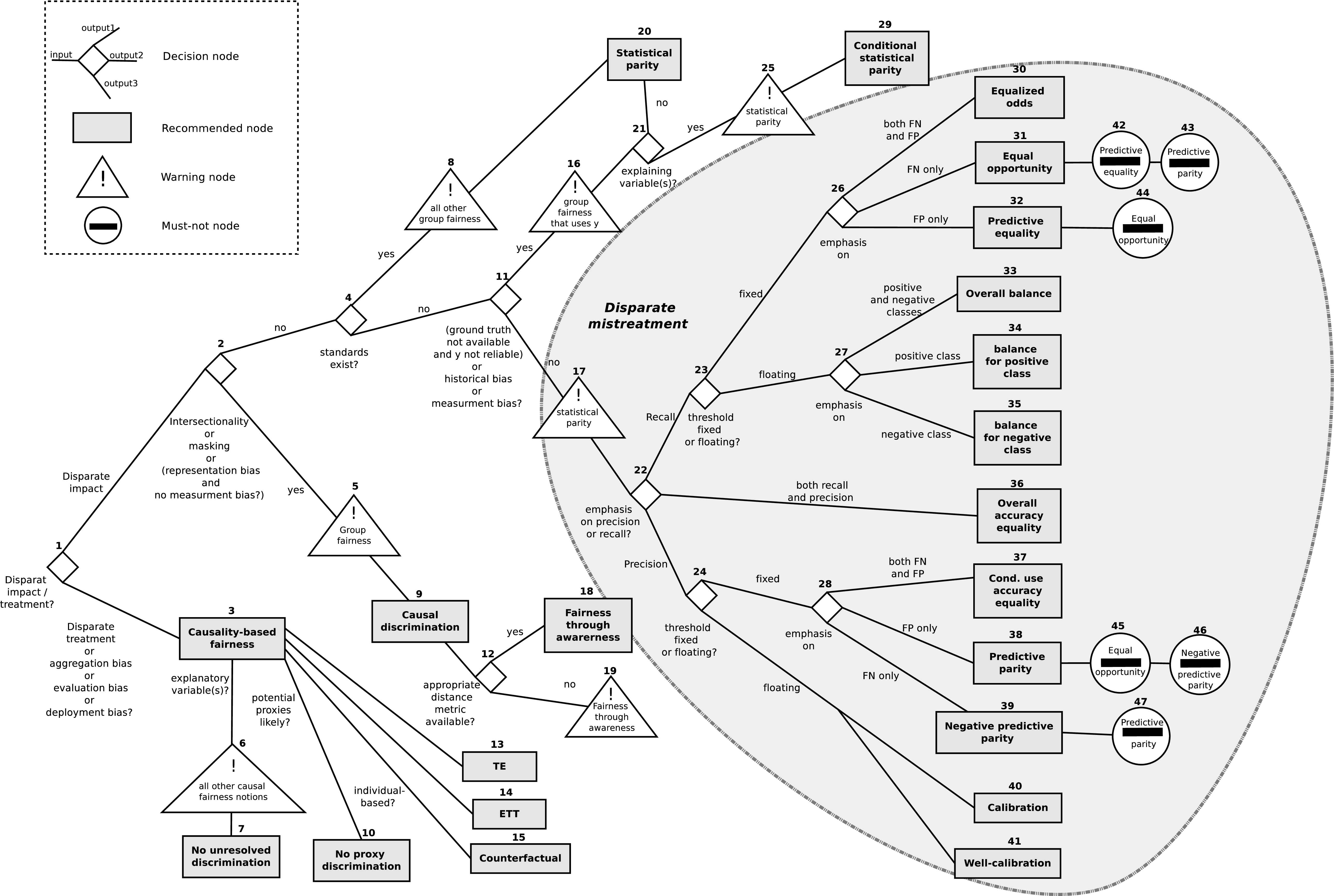}
    \caption{Fairness notions applicability decision diagram.}
     \label{fig:decisionDiagram}
\end{figure}
\end{landscape}

With the large number of fairness notions and the subtle resemblance between MLDM scenarios, deciding about which fairness notion to use is not a trivial task. More importantly, selecting and using a fairness notion in a scenario inappropriately may detect unfairness in an otherwise fair scenario, or the opposite, i.e., fail to identify unfairness in an unfair scenario.

One of the objectives of this survey is to systemize the selection procedure of fairness notions. This is achieved by identifying a set of fairness-related characteristics (Section~\ref{sec:criteria}) of the scenario at hand and then use them to recommend the most suitable fairness notion for that specific scenario. The proposed systemized selection procedure is illustrated in the decision diagram of Figure~\ref{fig:decisionDiagram}. The diagram is called ``decision diagram'' and not ``decision tree'' for the following reason. In typical decision trees, every leaf corresponds to a single decision, which is a fairness notion that \textit{should} be used. However, the diagram in Figure~\ref{fig:decisionDiagram} is designed such that every node indicates which notions are recommended, which notions \textcolor{black}{to be avoided}, and which notions must not be used. In addition, if a notion is not mentioned along the path, it means, it can be safely used.

The diagram is composed of four types of nodes:
\begin{itemize}
	\item \textbf{Decision node (diamond):} based on fairness-related characteristics (Section~\ref{sec:criteria}).
	\item \textbf{Recommended node (rectangle):} a leaf node indicating that the fairness notion is suitable to be used given all fairness-related characteristics in the path to that node.
	\item \textbf{Warning node (triangle):} indicates that the fairness notion(s) is/are not recommended in all the branch in the right of the node. This node can appear in the middle of the edge between two decision nodes. 
	\item \textbf{Must-not node (circle):} the fairness notion must not be used.
\end{itemize}

{\color{black}To illustrate how the diagram should be interpreted, consider the recommended node predictive parity (node 38). According to the diagram, predictive parity is recommended in the scenario where the legal framework is disparate impact (decision node 1), intersectionality and/or masking are unlikely (decision node 2), there is no evidence that representation bias is likely (decision node 2), standards do not exist (decision node 4), ground-truth is available or outcome $Y$ is reliable (decision node 11), historical and measurement bias are unlikely (decision node 11), fairness is more sensitive to precision rather than recall (decision node 22), the prediction threshold is typically fixed (decision node 24) and the emphasis is on false positives rather than false negatives (decision node 28). In that particular scenario, equal opportunity must not be used (must-not node 45) because fairness in this scenario is particularly sensitive to false positives, while equal opportunity is completely insensitive to false positives. Similarly, negative predictive parity must not be used (must-not node 46) as fairness is sensitive to precision rather than recall. The warning node 17 along the same path indicates that statistical parity is not suitable in this scenario. Finally, any fairness notion for which there is no a warning node or a must-not node along the path of the scenario can be used in this scenario. For instance, all individual fairness notions can be used.

As concrete example of situations where predictive parity (node 38) is recommended, consider the following. In situations when the outcome is influenced by the decision, some statistical quantities (e.g. FN, TN, etc.) are unlikely to be observed, and hence, any fairness notion that is defined in terms of those quantities is not suitable to use. For example, in real-world cases of loan-granting, a loan application which is predicted to be defaulting, will not be approved. Consequently, both negative statistics (true negative (TN) and false negative (FN)) will not be typically observed. Hence, fairness notions such as equalized odds and equality of opportunity cannot be used as they are defined in terms of TN and FN. In such cases, predictive parity (node 38) is recommended. 

\textit{\textbf{Node 1:}} Assessing fairness is very often performed in the context of a legal case where a plaintiff is filing a claim against a party that is using an MLDM. According to real-world legislation, in particular, the American anti-discrimination law, this can fall into one the two legal frameworks, namely, disparate impact and disparate treatment. If the plaintiff is filing the claim under the disparate impact framework, she can prove the liability of the defendant by using an observational group or individual fairness notion as the goal is to show that the practices and policies used by the defendant are facially neutral but have a disproportionately adverse impact on the protected class~\cite{barocas2016big}. If, however, the plaintiff is filing a claim under the disparate treatment framework, observational fairness notions are often not enough to prove the liability of the defendant as the goal is to show that the defendant has used the sensitive attribute to take the discriminatory decision. The recommended fairness notions in that case are causality-based (recommended node 3) since all of them are expressed in terms of the causal effect of the sensitive attribute on the prediction. 

\textit{\textbf{Node 2:}} As explained above, any unintentional type of bias can also be "orchestrated" intentionally by decision makers with prejudicial views. For instance, decision makers can purposefully bias the data collection step to ensure that the MLDM remains less favorable to protected classes. To reliably assess the bias in presence of such masking attempts, all group fairness notions should be avoided as they are defined in terms of statistics about the different sub-populations and hence can more easily be gamed by prejudicial decision makers. Intersectionality is similar to masking as both lead to a discrimination which is difficult to detect using statistical measures and consequently requires more fine-grained measures. Therefore individual fairness notions are recommended in presence of both criteria (nodes 9 and 18).

\textit{\textbf{Nodes 2, \textcolor{black}{3}, and 11:}} In case one or more sources of bias are suspected ahead of time (before assessing fairness), the information can help warn against the use of some fairness notions. If representation bias is likely, the performance (accuracy) of the MLDM on under-represented categories will often be worse. Such disparity in performance between groups may lead to unreliable fairness assessment in case a group fairness notion is used, in particular disparate mistreatment notions (grayed section of the diagram). In such case, individual fairness notions can assess fairness more reliably provided that measurement bias is not likely (node 2). A suspicion of historical or measurement bias means that the features (X) and/or the label (Y) are not reliable. All group fairness notions using the label Y (disparate mistreatment) as well as individual notions are not recommended in that case. Statistical parity is recommended in such situation. Finally, in presence of either aggregation, evaluation, or deployment bias, causality-based fairness notions are recommended. The reason is that the interventional and counterfactual quantities used in the definitions of these notions go beyond mere correlations and hence allow to assess fairness more reliably in presence of such bias. For instance, Coston et al.~\cite{coston2020counterfactual} propose counterfactual formulations of fairness metrics to properly account for the effect of intervention (decision) on the outcome. Such effect is a type of deployment bias. 

\textit{\textbf{Node 3:}} As discussed in Section~\ref{sec:causalNotions}, there are several notions that use causal reasoning to assess fairness. Counterfactual fairness is suitable in case a fine-grained assessment is required as the equality of Eq.~\ref{eq:counterfactual} conditions on all features (X). Counterfactual fairness, however, requires strong assumptions to be applicable in real scenarios (the availability of the full causal model including the latent variables distributions). Total effect (TE), effect of treatment on treated (ETT), and no proxy discrimination (nodes 13, 14 and 10), on the other hand, require a weaker assumption to be applicable, namely, the identifiability of the causal quantities used in their definitions. No proxy discrimination is recommended in presence of potential proxies, however, the identification of proxy variables requires a domain expertise of the application at hand. Finally, in case there are variables in the causal graph which are correlated with the sensitive attribute but in a manner that is accepted as nondiscriminatory, no unresolved discrimination is recommended while the remaining causal based fairness notions should be avoided. No unresolved discrimination is easier to apply in practice as it only needs the availability of the causal graph.

\textit{\textbf{Node 4:}} To reduce inequality and historical discrimination against sub-populations, in particular, minorities, some states and organizations resort to equality standards and regulations such as the laws enforced by the US Equal Employment Opportunity Commission~\cite{eeoc}. In presence of such standards, to be deemed fair, an MLDM should satisfy such standards. Consequently, all what matters for fairness assessment is the proportion of positive prediction across all groups which corresponds to statistical parity.  

\textit{\textbf{Node 17:}} If no standards/regulations exist (node 4) and either the ground truth exists or the outcome label Y is available (node 11), statistical parity is not recommended (node 17) as it can lead to misleading results such as detecting unfairness in an otherwise fair scenario or failing to identify fairness in an unfair scenario. For instance, in stop-and-frisk real world scenario applied in New York city \textcolor{black}{starting 1990}~\cite{bellin2014inverse}\footnote{Assuming the absence of measurement bias.}, the ground truth is available as by frisking an individual, a police officer can know with certainty the presence or no of illegal substance. In such case, one or several disparate mistreatment notions (nodes 30-41) are more suitable to assess fairness.

\textit{\textbf{Nodes 22-47:}} The bulk of Figure~\ref{fig:decisionDiagram} is dedicated for disparate mistreatment fairness notions and the criteria leading to each one of them. These notions define fairness in terms of the disparity of misclassification rates among the different groups in the population. Based on their definitions, selecting the most suitable notion to use depends on four citeria, namely, whether the emphasis is on precision or recall (node 22), whether the threshold is fixed or floating (nodes 23 and 24), whether the emphasis is on false negatives or false positives (nodes 26 and 28), and finally whether the emphasis is on the positive or negative class (node 27). As some notions focus only on either FP or FN (nodes 31, 32, 38, and 39), any notion that is insensitive to either FP or FN must not be used (nodes 42 - 47). 

The diagram may be misleading if it is interpreted very categorically. This occurs when a user of the diagram navigates it and ends up using the recommended fairness notion without considering other important elements specific to the scenario at hand. The diagram can be misleading also when it is not clear which branch to take in a decision node. For example, the question in decision node 22 (emphasis on precision or recall?) is difficult to answer categorically in several scenarios. The decision nodes 4, 21, 12, and even 2, are typically easier to navigate, but can be challenging to settle in a number of scenarios.  Moreover, in presence of measurement bias, the values of some features and even the outcome label may not be reliable which can make the diagram navigation more challenging. A potential solution would be to label one of the branches as default (to be followed when the answer is not clear), but this can, often result in a suboptimal decision. In summary, the diagram should be considered as guide and should never be used to supersede important elements specific to the scenario at hand.
  
\begin{landscape}
\begin{table}
\centering
\setstretch{1.0}
 \caption{\textcolor{black}{\scriptsize{Correspondence between Fairness notions and  the selection criteria: \textbf{C1}: disparate impact , \textbf{C2}: disparate treatment , \textbf{C3}: intersectionality/masking, \textbf{C4}: historical bias, \textbf{C5}: representational bias, \textbf{C6}: measurement bias, \textbf{C7}: aggregation/evaluation/deployment bias, \textbf{C8}: standards, \textbf{C9}: ground truth available, \textbf{C10}: $y$ not reliable, \textbf{C11}: explanatory variables, \textbf{C12}: precision, \textbf{C13}: recall, \textbf{C14}: FP, \textbf{C15}: FN, \textbf{C16}: causal graph available, \textbf{C17}: threshold floating.} \\ \scriptsize{Notation: \ding{51}: recommended, \danger: warning, \ding{55}: must not, $-$: insensitive.}}}
 \label{tab:notions_criteria}
 \setlength\extrarowheight{5pt} 
 \color{black}
\begin{tabular}{|l?B|B?B?B|B|B|B?B|B|B|B?B|B?B|B?B|B|}
\multicolumn{1}{c}{}& \multicolumn{2}{c}{\tiny{\textbf{Legal Frame}}}&  \multicolumn{1}{c}{}& \multicolumn{4}{c}{\tiny{\textbf{Suspected source of bias}}}& \multicolumn{1}{c}{} & \multicolumn{1}{c}{}& \multicolumn{1}{c}{}&\multicolumn{1}{c}{} & \multicolumn{2}{c}{\tiny{\textbf{Emphasis on}}}& \multicolumn{2}{c}{\tiny{\textbf{Emphasis on}}}&\multicolumn{1}{c}{} & \multicolumn{1}{c}{}  \\
 \hline
\backslashbox{\tiny{\textbf{Fairness notion}}}{\tiny{\textbf{Criterion}}}&  \tiny{\textbf{C1}} & \tiny{\textbf{C2}}& \tiny{\textbf{C3}} & \tiny{\textbf{C4}} & \tiny{\textbf{C5}} & \tiny{\textbf{C6}} & \tiny{\textbf{C7}} & \tiny{\textbf{C8}}& \tiny{\textbf{C9}} & \tiny{\textbf{C10}}& \tiny{\textbf{C11}}& \tiny{\textbf{C12}} & \tiny{\textbf{C13}} & \tiny{\textbf{C14}} & \tiny{\textbf{C15}} & \tiny{\textbf{C16}}& \tiny{\textbf{C17}}\\
\hline
\scriptsize{Statistical parity} & \footnotesize{\ding{51}}  & \footnotesize{\danger} &\footnotesize{\danger} & \footnotesize{\ding{51}} &\footnotesize{\danger} & \footnotesize{\danger}&\footnotesize{\danger} &\footnotesize{\ding{51}} &\footnotesize{\danger}& \footnotesize{\ding{51}} & \footnotesize{\danger}&$-$ &$-$ & $-$& $-$ & $-$&\footnotesize{\danger} \\

\scriptsize{Conditional statistical parity} & \footnotesize{\ding{51}}  & \footnotesize{\danger} &\footnotesize{\danger} & \footnotesize{\ding{51}} & \footnotesize{\danger}& \footnotesize{\danger}&\footnotesize{\danger} & $-$& \footnotesize{\danger}&\footnotesize{\ding{51}} &\footnotesize{\ding{51}} & $-$& $-$&$-$ & $-$&$-$& \footnotesize{\danger}\\
\scriptsize{Equalized odds} & \footnotesize{\ding{51}}  & \footnotesize{\danger} &\footnotesize{\danger} & \footnotesize{\danger} & \footnotesize{\danger}& \footnotesize{\danger}& \footnotesize{\danger}&$-$ & \footnotesize{\ding{51}}& \footnotesize{\danger}& \footnotesize{\danger}&\footnotesize{\danger} & \footnotesize{\ding{51}}& \footnotesize{\ding{51}}& \footnotesize{\ding{51}}&$-$ & \footnotesize{\danger}\\
\scriptsize{Equal opportunity} & \footnotesize{\ding{51}}  & \footnotesize{\danger} &\footnotesize{\danger} &\footnotesize{\danger}  & \footnotesize{\danger}& \footnotesize{\danger}& \footnotesize{\danger}&$-$ & \footnotesize{\ding{51}}& \footnotesize{\danger}&\footnotesize{\danger} &\footnotesize{\danger} &\footnotesize{\ding{51}} & \footnotesize{\ding{55}}& \footnotesize{\ding{51}}& $-$ & \footnotesize{\danger}\\
\scriptsize{Predictive equality} & \footnotesize{\ding{51}}  & \footnotesize{\danger} & \footnotesize{\danger}& \footnotesize{\danger} & \footnotesize{\danger}&\footnotesize{\danger} &\footnotesize{\danger} & $-$&\footnotesize{\ding{51}} & \footnotesize{\danger}&\footnotesize{\danger} &\footnotesize{\danger} &\footnotesize{\ding{51}} & \footnotesize{\ding{51}}& \footnotesize{\ding{51}}& $-$ &\footnotesize{\danger} \\
\scriptsize{Balance for positive class}  & \footnotesize{\ding{51}}  & \footnotesize{\danger} &\footnotesize{\danger} & \footnotesize{\danger} & \footnotesize{\danger}&\footnotesize{\danger} & \footnotesize{\danger}& $-$& \footnotesize{\ding{51}}&\footnotesize{\danger} &\footnotesize{\danger} & \footnotesize{\danger}&\footnotesize{\ding{51}} &\footnotesize{\ding{55}} & \footnotesize{\ding{51}}& $-$ & \footnotesize{\ding{51}}\\
\scriptsize{Balance for negative class}& \footnotesize{\ding{51}}  & \footnotesize{\danger} &\footnotesize{\danger} & \footnotesize{\danger} & \footnotesize{\danger}&\footnotesize{\danger} &\footnotesize{\danger} &$-$ &\footnotesize{\ding{51}} &\footnotesize{\danger} &\footnotesize{\danger} &\footnotesize{\danger} & \footnotesize{\ding{51}}& \footnotesize{\ding{51}}& \footnotesize{\ding{55}}& $-$ & \footnotesize{\ding{51}}\\
\scriptsize{Overall balance} & \footnotesize{\ding{51}}  & \footnotesize{\danger} & \footnotesize{\danger}& \footnotesize{\danger} & \footnotesize{\danger}& \footnotesize{\danger}&\footnotesize{\danger} &$-$ & \footnotesize{\ding{51}}& \footnotesize{\danger}& \footnotesize{\danger}& \footnotesize{\danger}&\footnotesize{\ding{51}} &\footnotesize{\ding{51}} &\footnotesize{\ding{51}}& $-$ & \footnotesize{\ding{51}}\\
\scriptsize{Conditional use acc. equality} & \footnotesize{\ding{51}}  & \footnotesize{\danger} & \footnotesize{\danger}& \footnotesize{\danger} &\footnotesize{\danger} &\footnotesize{\danger} &\footnotesize{\danger} &$-$ & \footnotesize{\ding{51}}&\footnotesize{\danger} & \footnotesize{\danger}& \footnotesize{\ding{51}}&\footnotesize{\danger} &\footnotesize{\ding{51}} &\footnotesize{\ding{51}} & $-$ & \footnotesize{\danger}\\
\scriptsize{Predictive parity} &  \footnotesize{\ding{51}} & \footnotesize{\danger} &\footnotesize{\danger} & \footnotesize{\danger} & \footnotesize{\danger}&\footnotesize{\danger} & \footnotesize{\danger}&$-$ & \footnotesize{\ding{51}}&\footnotesize{\danger} & \footnotesize{\danger}&\footnotesize{\ding{51}} &\footnotesize{\danger} & \footnotesize{\ding{51}}& \footnotesize{\ding{55}}& $-$ & \footnotesize{\danger}\\
\scriptsize{Negative predictive parity} & \footnotesize{\ding{51}}  & \footnotesize{\danger} & \footnotesize{\danger}& \footnotesize{\danger} &\footnotesize{\danger} &\footnotesize{\danger} & \footnotesize{\danger}&$-$ &\footnotesize{\ding{51}} & \footnotesize{\danger}&\footnotesize{\danger} & \footnotesize{\ding{51}}&\footnotesize{\danger} &\footnotesize{\ding{55}} & \footnotesize{\ding{51}}& $-$ & \footnotesize{\danger}\\
\scriptsize{Calibration} & \footnotesize{\ding{51}}  & \footnotesize{\danger} &\footnotesize{\danger} & \footnotesize{\danger} &\footnotesize{\danger} &\footnotesize{\danger} &\footnotesize{\danger} &$-$ & \footnotesize{\ding{51}}& \footnotesize{\danger}& \footnotesize{\danger}& \footnotesize{\ding{51}}& \footnotesize{\danger}& $-$&$-$ & $-$ & \footnotesize{\ding{51}}\\
\scriptsize{Well-calibration} & \footnotesize{\ding{51}}  & \footnotesize{\danger} &\footnotesize{\danger} & \footnotesize{\danger} &\footnotesize{\danger} &\footnotesize{\danger} &\footnotesize{\danger} &$-$ & \footnotesize{\ding{51}}& \footnotesize{\danger}&\footnotesize{\danger} & \footnotesize{\ding{51}}& \footnotesize{\danger}&$-$ &$-$ & $-$ & \footnotesize{\ding{51}}\\
\scriptsize{Overall accuracy equality} & \footnotesize{\ding{51}}   & \footnotesize{\danger} & \footnotesize{\danger}& \footnotesize{\danger} &\footnotesize{\danger} &\footnotesize{\danger} &\footnotesize{\danger} &$-$ &\footnotesize{\ding{51}} & \footnotesize{\danger}&\footnotesize{\danger} &\footnotesize{\ding{51}} &\footnotesize{\ding{51}} & \footnotesize{\ding{51}}&\footnotesize{\ding{51}} & $-$ & \footnotesize{\danger}\\
\scriptsize{Treatment equality} & \footnotesize{\ding{51}}  & \footnotesize{\danger} & \footnotesize{\danger}&\footnotesize{\danger}  & \footnotesize{\danger}&\footnotesize{\danger} & \footnotesize{\danger}& $-$& \footnotesize{\ding{51}}&\footnotesize{\danger} &\footnotesize{\danger} &$-$ & $-$&\footnotesize{\ding{51}} & \footnotesize{\ding{51}}& $-$ & $-$\\
\scriptsize{Total fairness} & \footnotesize{\ding{51}}  & \footnotesize{\danger} &\footnotesize{\danger} &  \footnotesize{\danger}&\footnotesize{\danger} & \footnotesize{\danger}&\footnotesize{\danger} &$-$ & \footnotesize{\ding{51}}& \footnotesize{\danger}& \footnotesize{\danger}& $-$&$-$ & \footnotesize{\ding{51}}& \footnotesize{\ding{51}}& $-$& \footnotesize{\danger}\\
\scriptsize{Causal discrimination} & \footnotesize{\ding{51}}  & \footnotesize{\danger} & \footnotesize{\ding{51}} & \footnotesize{\danger} & \footnotesize{\ding{51}}& \footnotesize{\danger}& $-$& $-$&\footnotesize{\ding{51}} &\footnotesize{\danger} & \footnotesize{\danger}&$-$ &$-$ & $-$&$-$ &$-$ & $-$\\
\scriptsize{Fairness through awareness}  & \footnotesize{\ding{51}} & \footnotesize{\danger}  & \footnotesize{\ding{51}}&  \footnotesize{\danger}&\footnotesize{\ding{51}} & \footnotesize{\danger}&\footnotesize{\danger} & $-$&\footnotesize{\ding{51}} &\footnotesize{\danger} &$-$ &$-$ & $-$& $-$& $-$&  $-$& $-$\\
\scriptsize{Total effect} & $-$  & \footnotesize{\ding{51}} & \footnotesize{\danger}& $-$ &$-$ & $-$& \footnotesize{\ding{51}}&$-$ &$-$ &$-$ &$-$ &$-$ & $-$& $-$&$-$ & \footnotesize{\ding{51}} & $-$\\
\scriptsize{Effect of treatment on treated} & $-$ & \footnotesize{\ding{51}} & \footnotesize{\danger}& $-$ & $-$&$-$ &\footnotesize{\ding{51}} & $-$& $-$& $-$& $-$& $-$& $-$&$-$ &$-$ & \footnotesize{\ding{51}} & $-$\\
\scriptsize{Counterfactual fairness} & $-$  & \footnotesize{\ding{51}} & \footnotesize{\ding{51}}& $-$ & \footnotesize{\ding{51}}&$-$ &\footnotesize{\ding{51}} & $-$&$-$ &$-$ &$-$ &$-$ &$-$ &$-$ &$-$ & \footnotesize{\ding{51}} & $-$\\
\scriptsize{No unresolved discrimination} & $-$  & \footnotesize{\ding{51}} &\footnotesize{\danger} & $-$ &$-$ & $-$&\footnotesize{\ding{51}} &$-$ &$-$ &$-$ & \footnotesize{\ding{51}}& $-$&$-$ &$-$ &$-$ & \footnotesize{\ding{51}} & $-$\\
\scriptsize{No proxy discrimination} &  $-$ & \footnotesize{\ding{51}} & \footnotesize{\danger}& $-$ & $-$& $-$& \footnotesize{\ding{51}}&$-$ &$-$ & $-$&$-$ &$-$ & $-$&$-$ & $-$ & \footnotesize{\ding{51}} & $-$\\
    \hline
\end{tabular}
\end{table}
\end{landscape}

Finally, Table~\ref{tab:notions_criteria} states explicitly the relationship between every selection criterion and every fairness notion. The table uses four symbols, namely, recommended (\ding{51}), warning (\danger), must-not (\ding{55}), and insensitive ($-$). Insensitive means that the choice of the fairness notion is independent of the selection criterion.}

\section{Conclusion}
With the increasingly large number of fairness notions considered in the relatively new field of fairness in ML, selecting a suitable notion for a given MLDM (machine learning decision making) becomes a non-trivial task. There are two contributing factors. First, the boundaries between the defined notions are increasingly fuzzy. Second, applying inappropriately a fairness notion may report discrimination in an otherwise fair scenario, or vice versa, fail to identify discrimination in an unfair scenario. This survey tries to address this problem by identifying fairness-related characteristics of the scenario at hand and then use them to recommend and/or discourage the use of specific fairness notions. The main contribution of this survey is to systemize the selection process based on a decision diagram.  Navigating the diagram will result in recommending and/or discouraging the use of fairness notions.

One of the main objectives of this survey is to bridge the gap between the real-world use case scenarios of automated (and generally unintentional) discrimination and the mostly technical tackling of the problem in the literature. Hence, the survey can be of particular interest to civil right activists, civil right associations, anti-discrimination law enforcement agencies, and practitioners in fields where automated decision making systems are increasingly used. 

More generally, in real-scenarios, there are still two important obstacles to address the unfairness problem in automated decision systems. First, the victims of such systems are, very often, members of minority groups with limited influence in the public sphere. Second, automated decision systems are geared towards efficiency (typically money) and to optimize profit, they are designed to sacrifice the outliers as tolerable collateral damage. After all, the system is benefiting most of the population (employers finding ideal candidates, banks giving loans to minimum risk borrowers, a society with recidivists locked in prisons, etc.).

\section*{Acknowledgement}
The work of Catuscia Palamidessi was supported by the European Research Council (ERC) under the European Union’s Horizon 2020 research and innovation programme. Grant agreement 835294. 


\appendix

\section{Counterfactual probability computation using the three-step procedure}
\label{ap:abductionComputation}

\begin{figure}[!ht]
\centering
\includegraphics [scale=0.3] {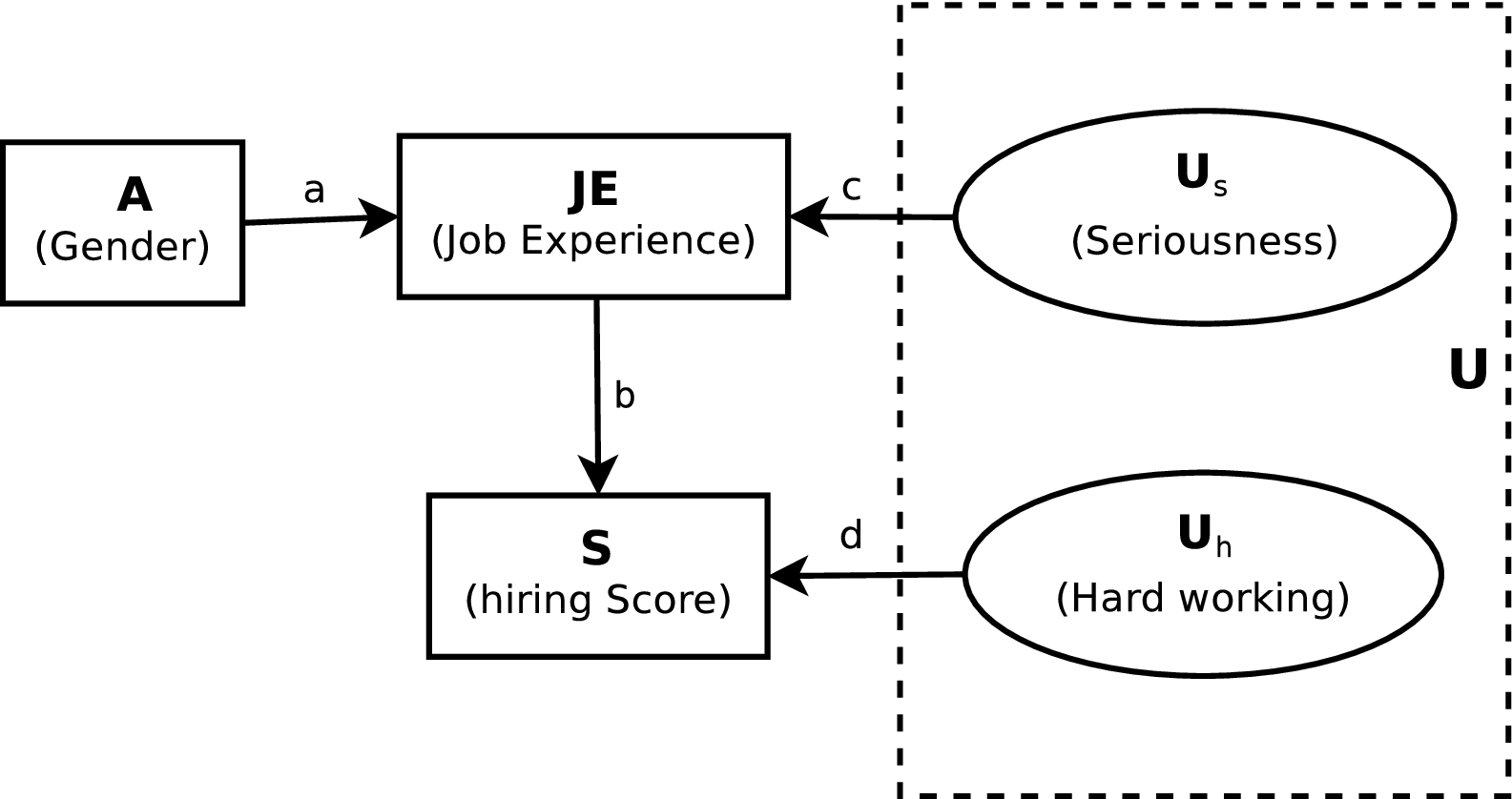}
\caption{A simple deterministic causal graph for the hiring example.}
\label{fig:simpleGraph}     
\end{figure}

The probability of the counterfactual realization $ P(\hat{Y}_{A \leftarrow a_1}  \mid X =  x, A =  a_0)$ is computed using the following three-steps process~\citep{pearl2000book}:
	\begin{enumerate}
		\item \textbf{Abduction}: update the probability $P(U=u)$ given the evidence to obtain: $P(U=u \mid X=x, A=a_0)$.
		\item \textbf{Action}: set the sensitive attribute value $A$ to $a_1$ and update all structural functions of the causal graph accordingly.
		\item \textbf{Prediction}: compute the outcome $(\hat{Y})$ value using the updated probability $P(U \mid X=x, A=a_0)$ and structural functions.
\end{enumerate}

To illustrate how counterfactual quantities are computed, consider the simplified deterministic version of the hiring example in Figure~\ref{fig:simpleGraph}. For simplicity, the hiring score variable $S$ depends on the observable variable $JE$ representing job experience and the exogenous variable $U_{h}$ representing how hard working the candidate is. The variable $JE$ in turn depends on the observable sensitive variable $A$ representing the gender (male or female) and the exogenous variable $U_{s}$ representing the seriousness of the candidate. The causal graph in Figure~\ref{fig:simpleGraph} is represented by the two following equations:
\begin{eqnarray}
	JE &=& a.A + c.U_{s} \label{eq:je}\\
	S  &=& b.JE + d.U_{h} \label{eq:s}
\end{eqnarray}

For simplicity of the illustration, assume that both $U$ ($U_s$ and $U_h$) variables are independent and all the parameters of the model (Eq.~\ref{eq:je} and~\ref{eq:s}) are known. Assume that the values of the coefficients are given as follows:
$$ a=0.1, \quad b=0.7, \quad c=0.9, \quad d=0.3$$
Given this causal model, consider a candidate John who is male ($A^{John}=1$), with the normalized\footnote{To keep the computation simple, all variable values are normalized between $0$ and $1$.} job education level $JE^{John}=0.6$ and a predicted score $\hat{S}^{John}=0.55$. Assessing the fairness of the hiring score prediction with respect to gender is achieved through answering the following question: \textit{what would John's hiring score have been had he was of opposite gender (female)?} This corresponds to the hiring score of John in the counterfactual world where John is a female ($\hat{S}^{John}_{A \leftarrow 0}$). To compute this quantity, the three-steps process above is used, namely, abduction, action, and prediction. 

The abduction step consists in using the evidence ($A^{John}=1, JE^{John}=0.6, \hat{S}^{John}=0.55$) to identify the specific characteristics of $John$, namely, his level of seriousness and hard working ($U_{s}$ and $U_{h}$)\footnote{Since this example is deterministic, every individual is characterized by a unique assignment for exogenous variables $U_{s}$ and $U_{h}$. In typical (non-deterministic) scenarios, every individual is assigned a probability distribution over the exogenous variables.} as follows:
\begin{eqnarray}
	U_{s}^{John} &=& \frac{JE^{John} - a.A^{John}}{c} \nonumber  \\
		     &=& \frac{5}{9} \nonumber \\
		     &&\\
		     U_{h}^{John}  &=& \frac{\hat{S}^{John} - b.JE^{John}}{d}  \nonumber\\
		     		   &=& \frac{13}{30} \nonumber
\end{eqnarray}

The second step consists in setting the sensitive attribute $A^{John}$ to the opposite gender ($0$) and updating all equations of the model. This consists in replacing the variable $A$ in Eq.~\ref{eq:je} by $0$.  

The third step consists in the prediction, that is computing $\hat{S}_{A \leftarrow 0}$ in the counterfactual world. This requires the computation of $JE^{John}_{A \leftarrow 0}$, that is, the job experience of John in a world where John is a female.

\begin{eqnarray}
	JE^{John}_{A \leftarrow 0} &=& a.0 + c.U^{John}_{s} \nonumber \\
	& = & 0.5 \nonumber \\
	&& \\
	\hat{S}^{John}_{A \leftarrow 0} &=& b.JE^{John}_{A \leftarrow 0} + d.U^{John}_{h} \nonumber \\
	& = & 0.48
\end{eqnarray}

Hence, the hiring score of John had he was female is $\hat{S}^{John}_{A \leftarrow 0}=\textcolor{black}{0.48}$ which is considered a violation of counterfactual fairness as the predicted hiring score of John in the original world is $\hat{S}^{John}=0.55$.

Consider now a female candidate Marie ($A^{Marie}=0$), with the a job education level $JE^{Marie}=0.61$ and a predicted score $\hat{S}^{Marie}=0.65$. The question to investigate is now: \textit{what would Marie's hiring score have been had she was male?} This boils down to computing $\hat{S}^{Marie}_{A \leftarrow 1}$ and comparing it with $\hat{S}^{Marie}=0.65$. Applying the three-steps process:

Abduction:
\begin{eqnarray}
	U_{s}^{Marie} &=& \frac{JE^{Marie} - a.A^{Marie}}{c} \nonumber \\
		     &=& \frac{61}{90} \nonumber \\
		     &&\\
		     U_{h}^{Marie}  &=& \frac{\hat{S}^{Marie} - b.JE^{Marie}}{d} \nonumber\\
		     		   &=& \frac{223}{30} \nonumber
\end{eqnarray}

Action: replacing the variable $A$ in Eq.~\ref{eq:je} by $1$.

Prediction:
\begin{eqnarray}
	JE^{Marie}_{A \leftarrow 1} &=& a.1 + c.U^{Marie}_{s} \nonumber \\
	& = & 0.71 \nonumber \\
	&& \\
	\hat{S}^{Marie}_{A \leftarrow 1} &=& b.JE^{Marie}_{A \leftarrow 1} + d.U^{Marie}_{h} \nonumber \\
	& = & 0.72
\end{eqnarray}

$\hat{S}^{Marie}_{A \leftarrow 1}= 0.72 > \hat{S}^{Marie}=0.65$ is another violation for counterfactual fairness.

\bibliography{Fairness_Applicability}
\bibliographystyle{abbrv}

\end{document}